\documentclass{article}

\usepackage{arxiv}
\usepackage[square,numbers]{natbib}
\usepackage[utf8]{inputenc} 
\usepackage[T1]{fontenc}    
\usepackage{hyperref}       
\usepackage{url}            
\usepackage{booktabs}       
\usepackage{amsfonts}       
\usepackage{nicefrac}       
\usepackage{microtype}      
\usepackage{lipsum}		
\usepackage{graphicx}
\usepackage{doi}
\usepackage{tabularx}
\usepackage{graphicx}
\graphicspath{{Figures_v2/}}     
\usepackage{flafter}   
\usepackage[section]{placeins} 
\usepackage{subcaption}
\usepackage{amsmath}
\usepackage{mdframed}
\usepackage{longtable}
\newcolumntype{P}[1]{>{\raggedright\arraybackslash}p{#1}}
\usepackage{comment}
\usepackage{csquotes}
\usepackage[table]{xcolor}
\usepackage{multirow}
\usepackage{array}
\usepackage{listings}
\usepackage{soul}
\usepackage{xcolor}
\usepackage{fancyvrb}

\usepackage{enumitem}
\setlist[itemize]{noitemsep, topsep=0pt, parsep=0pt, partopsep=0pt}

\usepackage{titlesec}
\titleformat{\paragraph}[block]
  {\normalfont\normalsize\bfseries}
  {\thesubsubsection.\arabic{paragraph}}{1em}{}
\titlespacing*{\paragraph}{0pt}{1.25ex plus .2ex minus .2ex}{0.75ex}



\usepackage{booktabs,tabularx,makecell,caption}
\usepackage{lmodern}          
\usepackage[T1]{fontenc}
\usepackage{microtype}
\usepackage{enumitem}         
\usepackage[most]{tcolorbox}  
\tcbuselibrary{breakable,skins}
\DeclareFontShape{T1}{lmr}{bx}{sc}{<-> ssub * lmr/m/sc}{}
\usepackage{pgfplots}
\pgfplotsset{compat=1.18}

\usepackage{tabularx}
\usepackage{booktabs,longtable}
\usepackage{placeins} 
\usepackage{float}

\usepackage{booktabs,tabularx,array,graphicx,multirow}
\renewcommand{\arraystretch}{1.01}  

\usepackage{graphicx,array,multirow,booktabs}
\usepackage{graphicx,multirow,booktabs,needspace}
\usepackage{booktabs,longtable,multirow,graphicx,adjustbox}

\newcommand{\ptblockstart}{\Needspace{9\baselineskip}}
\newcommand{\ptblocksep}{\cmidrule(lr){3-7}}

\newcommand{\setPT}[1]{\gdef\CurrentPT{#1}}  
\newcommand{\setTK}[1]{\gdef\CurrentTK{#1}}   

\usepackage{booktabs,longtable,graphicx,adjustbox,array}
\newcommand{\rotv}[1]{\rotatebox[origin=c]{90}{\strut #1}}

\renewcommand{\arraystretch}{1.05}

\usepackage{longtable,booktabs,makecell,array,needspace}

\newcolumntype{C}[1]{>{\centering\arraybackslash}p{#1}}
\usepackage{needspace} 

\newtcolorbox{promptbox}[1][]{
  enhanced,
  breakable,                   
  colback=black!3,             
  colframe=black!35,           
  boxrule=0.35pt,              
  arc=1pt,                     
  top=6pt,bottom=6pt,left=8pt,right=8pt,
  coltitle=black,
  fonttitle=\bfseries,
  title=#1,
  borderline west={1.25pt}{0pt}{black!35}, 
}

\usepackage{mdframed}
\usepackage{ragged2e}   
\usepackage{microtype}  

\title{Evaluating Prompting Strategies and Large Language Models in Systematic Literature Review Screening: Relevance and Task-Stage Classification}

\author{\And\And\And
	\href{https://orcid.org/0000-0001-5376-8986}{\includegraphics[scale=0.06]{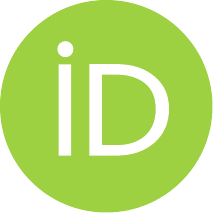}\hspace{1mm}Binglan Han\thanks{Corresponding author: b.han1@massey.ac.nz}}\\
	School of Mathematical and Computational Sciences\\
	Massey University\\
	Albany, New Zealand \\
	\And
 \And \And\And
	\href{https://orcid.org/0000-0002-9124-2536}{\includegraphics[scale=0.06]{orcid.pdf}\hspace{1mm}Anuradha Mathrani} \\
	School of Mathematical and Computational Sciences\\
	Massey University\\
	Albany, New Zealand \\
	\And
\href{https://orcid.org/0000-0001-9416-1435}{\includegraphics[scale=0.06]{orcid.pdf}\hspace{1mm} Teo Susnjak} \\
	School of Mathematical and Computational Sciences\\
	Massey University\\
	Albany, New Zealand \\
}

\renewcommand{\shorttitle}{Evaluating Prompting Strategies and LLMs in SLR Screening}

\begin{document}
\maketitle
\begin{abstract}
This study quantifies how prompting strategies interact with large language models (LLMs) to automate the screening stage of systematic literature reviews (SLRs). We evaluate six LLMs (GPT-4o, GPT-4o-mini, DeepSeek-Chat-V3, Gemini-2.5-Flash, Claude-3.5-Haiku, Llama-4-Maverick) under five prompt types (zero-shot, few-shot, chain-of-thought (CoT), CoT-few-shot, self-reflection) across relevance classification and six Level-2 tasks, using accuracy, precision, recall, and F1. Results show pronounced model–prompt interaction effects: CoT-few-shot yields the most reliable precision–recall balance; zero-shot maximizes recall for high-sensitivity passes; and self-reflection underperforms due to over-inclusivity and instability across models. GPT-4o and DeepSeek provide robust overall performance, while GPT-4o-mini performs competitively at a substantially lower dollar cost. A cost–performance analysis for relevance classification (per 1,000 abstracts) reveals large absolute differences among model–prompt pairings; GPT-4o-Mini remains low-cost across prompts, and structured prompts (CoT/CoT-few-shot) on GPT-4o-Mini offer attractive F1 at a small incremental cost. We recommend a staged workflow that (1) deploys low-cost models with structured prompts for first-pass screening and (2) escalates only borderline cases to higher-capacity models. These findings highlight LLMs' uneven but promising potential to automate literature screening. By systematically analysing prompt-model interactions, we provide a comparative benchmark and practical guidance for task-adaptive LLM deployment.
\end{abstract}


\section{Introduction}
Systematic literature reviews (SLRs) are a foundational method used in evidence-based research to provide transparent and reproducible syntheses of existing knowledge. The SLR process typically unfolds in five stages: (1) searching for relevant studies, (2) screening titles and abstracts to assess eligibility, (3) retrieving information from full texts, (4) synthesizing findings, and (5) writing results in a structured manner \cite{han2024automating}. Among these, the screening stage is especially demanding, as researchers must evaluate many publications precisely and consistently. Indeed, the rapid growth of scientific output has made manual screening increasingly difficult, which has in turn limited SLRs' scalability and timeliness \cite{omara2015using, borah2017analysis, marshall2019toward}.

Advances in artificial intelligence, particularly LLMs, present new opportunities to improve the screening stage of the SLR process \cite{li2024evaluating}. LLMs excel at natural language understanding, reasoning, and classification, which are capabilities that align with screening demands \cite{issaiy2024methodological}. Early studies have already demonstrated LLMs' potential to automate abstract screening by reducing manual workloads while maintaining accuracy \cite{nori2023capabilities, huang2023can}. By leveraging LLMs, researchers can likely increase screening efficiency while also scaling their evidence syntheses more effectively \cite{sanghera2025high}.

To best harness LLMs for SLR tasks, prompt engineering (deliberately designing inputs to steer model behaviour) is critical. Prompting strategies range from simple formats (zero-shot, few-shot) to more advanced methods (chain-of-thought, self-reflection); they aim to improve reasoning and calibration by eliciting intermediate steps or self-evaluation (\cite{wei2022chain, madaan2023selfrefine}). Although researchers have explored LLMs across several SLR tasks, recent work has focused most on automation of title/abstract screening \cite{issaiy2024methodological,syriani2024screening, cao2024prompting, li2024evaluating}. Within automation of title/abstract screening, one key gap remains: a factorial, cross-model evaluation that jointly varies prompt type, LLM family/version, and eligibility-criteria formulation to quantify interaction effects on accuracy, precision–recall balance, and screening efficiency in SLR automation.

As such, this study comparatively evaluates prompting strategies for literature screening across multiple LLMs. We assess zero-shot, few-shot, CoT, and self-reflection prompting with six state-of-the-art models: GPT-4o \cite{OpenAI2025GPT4o}, GPT-4o-Mini \cite{OpenAI2025GPT4oMini}, DeepSeek-Chat-V3 \cite{DeepSeek2025V3}, Llama-4-Maverick \cite{MetaAI2025Llama4Maverick}, Gemini-2.5-Flash \cite{Google2025Gemini25Flash}, and Claude-3.5-Haiku \cite{Anthropic2024Claude35Haiku}. By systematically analysing model-prompt interactions, we identify performance trade-offs, optimal configurations, and practical strategies to integrate LLMs into SLR workflows.

This study makes four key contributions to the automation of literature screening:
\begin{itemize}[noitemsep, topsep=0pt]
\setlength\itemsep{4pt}      
\item \textbf{Systematically evaluating prompting strategies: }Comprehensively assessing how different prompting approaches influence LLM performance in automated screening.
  \setlength\itemsep{4pt}   
\item \textbf{Benchmarking state-of-the-art LLMs: }Providing comparative evidence of leading models' strengths, weaknesses, and consistencies/inconsistencies in classifying systematic review tasks.
  \setlength\itemsep{4pt}   
\item \textbf{Identifying effective model-prompt configurations: }Offering practical guidance for LLM deployment in screening workflows by pinpointing model-prompt combinations that yield robust, balanced performance.
  \setlength\itemsep{4pt}   
\item \textbf{Advancing methodological understanding through evidence: }Providing evidence-based insight into multiple dimensions of literature screening automation by examining interactions among prompting strategies, LLMs, and classification tasks.
\end{itemize}\vspace{-\baselineskip} 

\section{Literature Review}

\subsection{Automating Literature Screening}
SLRs are scientific researchers' preferred means of synthesizing evidence; however, they remain very resource-intensive, particularly at the literature-screening stage. In screening literature, researchers assess up to thousands of retrieved citations against predefined inclusion and exclusion criteria. Multiple reviewers often must double-screen to ensure rigor, thus expending substantial time and effort before analysis can even commence \cite{omara2015using, borah2017analysis}. Consequently, academics rely increasingly on automation to reduce workload and enhance methodological rigor, transparency, and reproducibility.

Early work used supervised machine learning (ML) methods (support vector machines, naïve Bayes, and logistic regression) to classify abstracts \cite{wallace2010semi,thomas2011applications}. Relying on bag-of-words or TF–IDF features, these models required large labelled datasets and struggled with semantic refinement as well as cross-domain generalization. Later, semi-supervised and active learning approaches reduced labelling demands by prioritizing informative records \cite{cohen2006reducing, miwa2014reducing}. Active learning is now central to screening, with tools like ASReview reducing workload up to 70\% while maintaining recall \cite{van2021open}. Advances include (1) stopping heuristics (like the SAFE procedure) that balance efficiency with comprehensiveness \cite{boetje2024safe} and (2) ensemble methods that improve stability with classifier combinations \cite{marshall2018machine}. In parallel, platforms like Rayyan, EPPI-Reviewer, SWIFT-Active Screener, and Research Screener embed ML-driven prioritization to focus reviewer effort \cite{stansfield2021applying, chai2021research}. Benchmarking efforts (like the CLEF eHealth 2019 Technology-Assisted Review task) also establish frameworks for assessing generalizability \cite{kanoulas2019clef}. More recently, deep learning and transformer architectures have improved semantic modelling, though reproducibility and cross-domain transfer remain challenging \cite{bannach2021technological}.

LLM application constitutes the most significant recent shift in research. Exploratory studies using GPT-3.5, GPT-4, and related models show strong recall but variable precision, making such models appropriate triage aids that still cannot replace dual human screening \cite{dennstadt2024title, oami2024performance}. Hybrid strategies that combine LLMs with active learning or re-ranking approaches yield efficiency gains while preserving recall safeguards \cite{landschaft2024implementation, matsui2024three}. These findings point toward human-in-the-loop workflows, where AI supports prioritization and filtering while humans oversee final inclusion decisions.

\subsection{Language Models Used in Systematic Literature Reviews}
Transformer-based architecture (most notably BERT) has reshaped natural language processing by enabling contextualized word representations and deeper semantic modelling \cite{devlin2019bert}. Domain-specific variants like BioBERT \cite{lee2020biobert} and SciBERT \cite{beltagy2019scibert} extend these gains to biomedical and scientific corpora to improve screening and classification \cite{du2024machine, ambalavanan2020using}. This foundation facilitates SLR automation by showcasing transformers' capacity to parse complex abstracts and study reports. LLMs (like GPT-3, GPT-4, Claude, Gemini, and LLaMA) have since introduced broad generalizations with minimal fine-tuning, which allow zero-shot and few-shot applications where labelled data are scarce \cite{brown2020language, openai2023gpt4, anthropic2024claude, deepmind2024gemini, touvron2023llama}. Early evaluation reports their capacity to conduct screening, retrieval, and synthesis tasks framed as classification or question answering \cite{khraisha2024can, nori2023capabilities, sujau2025accelerating}. Prospective studies also report strong sensitivity but variable precision, again indicating that LLMs work well as prioritization aids but cannot replace dual human reviewers \cite{dennstadt2024title, oami2024performance}. Furthermore, hybrid designs embedding LLMs into active learning or re-ranking pipelines improve efficiency while safeguarding recall \cite{landschaft2024implementation, matsui2024three}.

LLMs also contribute to broader SLR automation (for instance, information extraction, synthesis, and drafting) \cite{guo2023automated, bandyopadhyay2024automating, susnjak2025automating}. Retrieval-augmented generation (RAG) enhances domain specificity and reduces hallucination, thereby promoting reproducibility \cite{ali2024automated, han2024automating}. However, performance depends on prompt design, model selection, and choice of evaluation metric; thus, prompt engineering and human oversight remain critical. In sum, while LLMs mark a major advancement in SLR automation, accuracy, transparency, and rigor still necessitate human-in-the-loop oversight. Full automation will demand further development, systematic testing, and validation. For now, LLMs serve best as intelligent assistants that accelerate screening and synthesis; indeed, they complement but cannot replace human judgment.

\subsection{Prompt Engineering for LLMs and Its Applications in SLR Automation}
Prompt engineering involves deliberately designing inputs to elicit accurate and contextually appropriate outputs; across tasks, it remains key to enhanced LLM performance \cite{reynolds2021prompt}. Once considered an ad-hoc practice, prompt engineering has become a structured field with theoretical foundations, design principles, and empirical validation \cite{white2023prompt}. Task framing, input format, and instruction specificity can significantly change LLM performance, so SLR automation relies heavily on prompt engineering. Basic strategies include (1) zero-shot prompting, where models complete tasks without examples, and (2) few-shot prompting, where limited exemplars guide output \cite{brown2020language}. These methods showcase LLMs' generalization capabilities, so they remain widely used in classification, summarization, and question answering. More advanced techniques extend model reasoning: (1) chain-of-thought (CoT) prompting elicits a step-by-step explanation that improves performance on an arithmetic, common-sense, or logical task \cite{wei2022chain, kojima2022large}, while (2) self-reflection prompting enables iterative critique and revision that makes complex problem solving more robust \cite{madaan2023selfrefine, liu2023selfreflection}. Such approaches move LLMs beyond surface-level pattern recognition toward structured reasoning, which is essential for evidence synthesis.

Although early research largely addressed domains like mathematics, common-sense reasoning, and programming \cite{zhou2022least, kojima2022large}, recent studies have applied prompt engineering to SLR workflows, particularly in the screening stage. These studies show that three levers shape abstract LLM screening: (1) prompting strategy, (2) model selection, and (3) expression of eligibility criteria. Early single-model work with GPT-3.5 used PICOS-anchored, zero-shot prompts before applying a relevance threshold to protect recall \cite{issaiy2024methodological}. In software-engineering corpora, zero-shot prompts that clearly list exclusion criteria outperform few-shot variants and exhibit improved specificity \cite{syriani2024screening}. Broader evaluations across multiple models converge on structured templates (either (a) criterion-by-criterion Boolean checks in a rigid output format or (b) framework-style reasoning) with deterministic decoding \cite{cao2024prompting, li2024evaluating}. Large multi-model and ensemble studies further demonstrate that small changes to bias wording in otherwise identical prompts shift the recall–precision balance (each model has a different optimal bias). Ensembling can secure very high sensitivity with workable precision \cite{sanghera2025high}. Live deployments report high recall using exclusion-first, PICO-grounded Boolean prompts with tightly controlled generation settings \cite{homiar2025development}. A cross-domain assessment also highlights that the wording and granularity of inclusion and exclusion criteria can change results as much as prompt style or model upgrades \cite{delgado2025transforming}. Moreover, systems with added retrieval components illustrate how topic and corpus characteristics interact with prompt structure and model selection while also motivating recall-protective decision rules \cite{chen2025medragent}.

\subsection{Research Questions}
Although LLMs have been explored in the context of screening automation, few studies systematically compare prompting strategies across models and screening criteria. Consequently, an important question remains: how do different model–prompt combinations influence performance, as assessed under standard evaluation metrics? Importantly, practical deployment requires accuracy but remains subject to operational constraints: screening typically spans tens of thousands of abstracts, token-expanding prompts materially increase spend and latency, and model choice can vary costs by several orders of magnitude. We address these problems by systematically assessing several prompting strategies and LLMs to provide empirical evidence and actionable guidance. Therefore, we ask:

\textbf{RQ1:} How do prompting strategies and LLM choice individually and jointly affect screening performance across standard metrics (accuracy, precision, recall, F1) and screening criteria?

\textbf{RQ2:} Given the scale and budget sensitivity of real-world screening, what are different model–prompt combinations' cost–performance implications for large-scale use, and which configurations deliver the best balance between accuracy and operational efficiency?

\section{Methodology}
\begin{figure}[!htbp]
  \centering
  \includegraphics[width=1.0\textwidth]{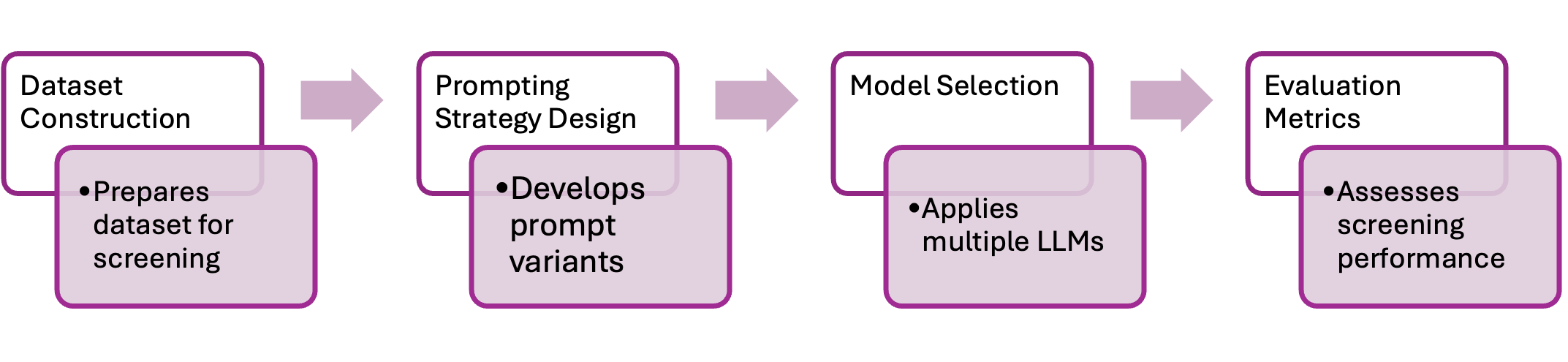}
  \caption{Workflow of Research Methodology}
  \label{fig:researchWorkflow}
\end{figure}

We examine how different LLMs perform with different prompting strategies in literature-screening automation. We structure our research (Figure \ref{fig:researchWorkflow}) around four main components: (1) dataset construction, (2) prompting strategy design, (3) model selection, and (4) evaluation metrics. The following subsections describe each component in detail.

\subsection{Dataset Construction}
To enable a comprehensive evaluation, we curated a dataset of academic publications on SLR automation. We retrieved papers published between 2014 and 2024 from major databases (Scopus, ScienceDirect, ACM Digital Library, IEEE Xplore, Web of Science, Semantic Scholar), using APIs where available. Search queries targeted terms such as ``literature review automation'' and ``systematic review automation''. Our initial search returned over 5,000 records. After removing duplicates and excluding non-research items, we were left with 1,376 unique studies for manual screening.

\begin{figure}[!htbp]
  \centering
  \includegraphics[width=1.0\textwidth]{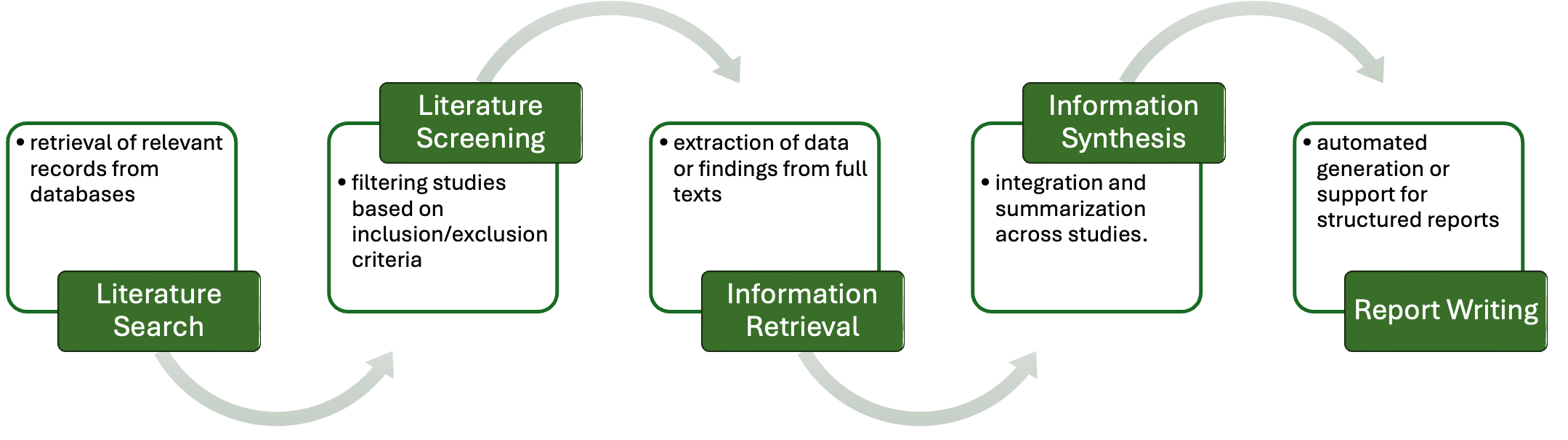}
  \caption{Stages of SLR Process}
  \label{fig:SLRStages}
\end{figure}

\begin{figure}[!htbp]
  \centering
  \includegraphics[width=1.0\textwidth]{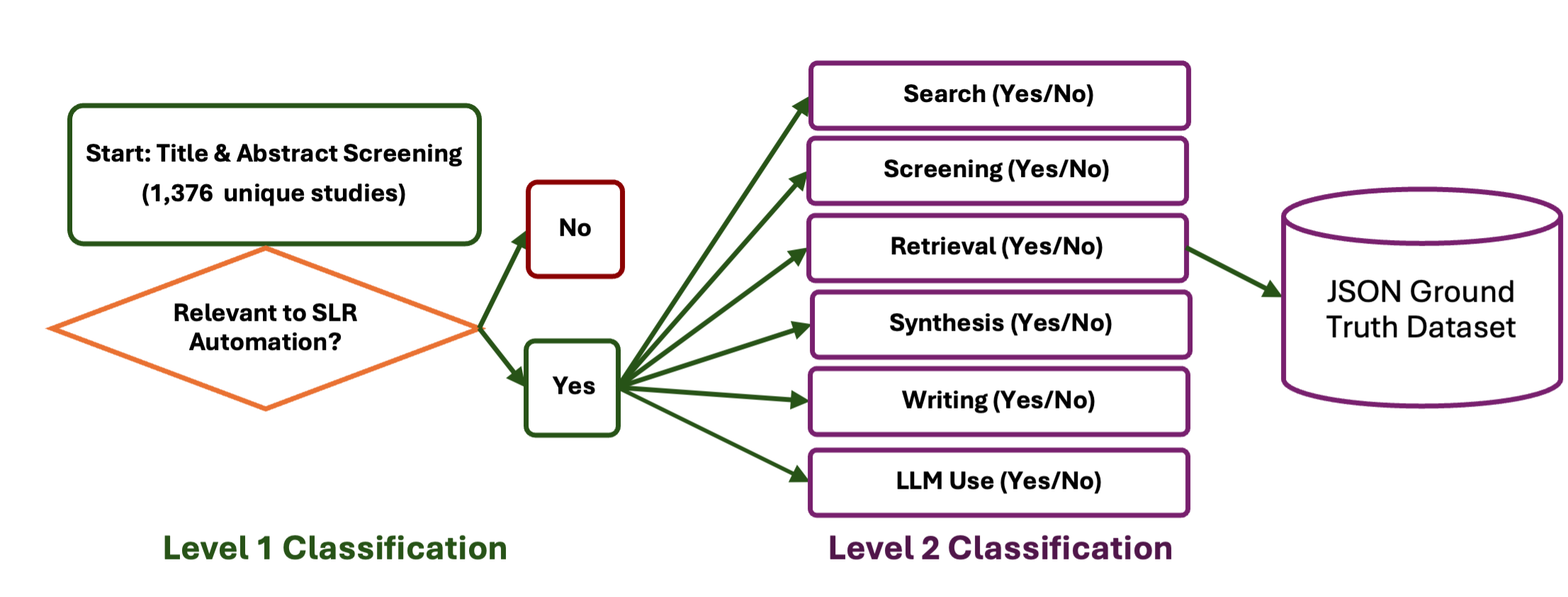}
  \caption{Screening Process Flowchart}
  \label{fig:screeningProcess}
\end{figure}

We screened titles and abstracts for eligibility; studies were eligible if they addressed automation of any SLR stage (searching, screening, retrieval, synthesis, report writing) (Figure \ref{fig:SLRStages}). We then annotated studies for LLM use. As shown in Figure \ref{fig:screeningProcess}, screening involved two levels of classification: \textbf{(1)} overall relevance to SLR automation \textit{(yes/no)} and \textbf{(2)} task-specific annotation (searching, screening, retrieval, synthesis, report writing) and detection of LLM use \textit{(yes/no)}. We formalized all labelled records into a structured JSON dataset that established a ground truth. To ensure reliable annotation, two researchers with LLM and systematic-review experience independently labelled all records, resolving disagreements by consensus. This produced a high-quality, expert-validated dataset as our foundation for LLM performance evaluation.

\begin{figure}[!htbp]
  \centering
  \includegraphics[width=0.7\textwidth]{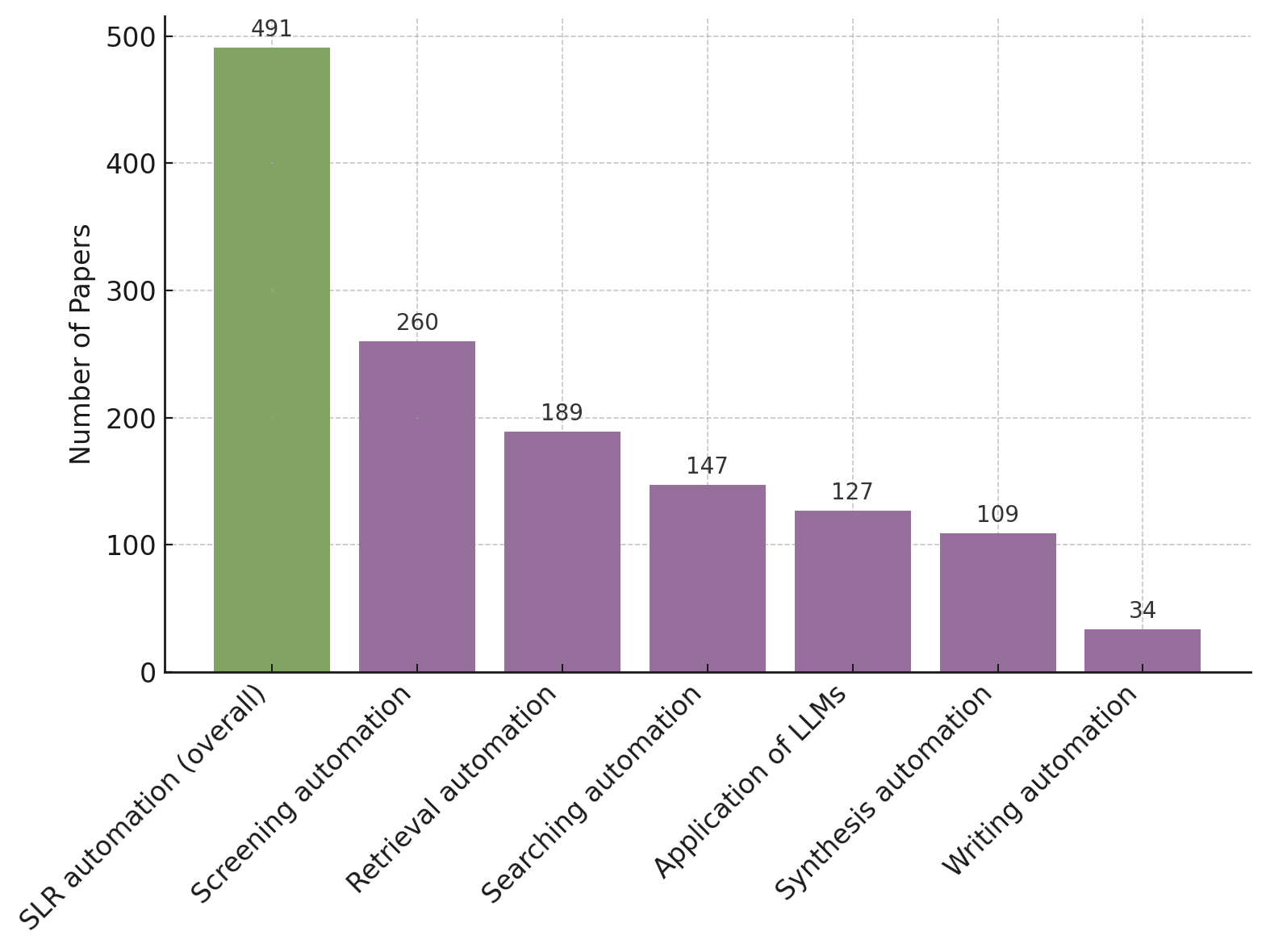}
  \caption{Distribution of Studies Relevant to SLR Automation Tasks}
  \label{fig:distributionStudies}
\end{figure}

As shown in Figure \ref{fig:distributionStudies}, of the 1,376 screened studies, we identified 491 (36.2\%) as relevant. Screening automation accounted for the largest share (260, 53\%), followed by retrieval (189, 38\%), searching (147, 30\%), synthesis (109, 22\%), and finally writing (34, 7\%). These categories were not mutually exclusive, as many studies addressed multiple SLR stages. The distribution above reflects task complexity: labour-intensive processes (like screening) received greater attention, while conceptually demanding tasks (like synthesis and writing) received less attention. Notably, 127 studies (26\%) explicitly examined LLM applications; this is a substantial proportion, given LLMs' recent emergence.

\subsection{Prompting Strategy Design}
To systematically assess input design's impact on LLM performance, we evaluated five prompting strategies that represent the primary paradigms used in both practice and research:

\begin{itemize}[noitemsep, topsep=0pt]
\setlength\itemsep{4pt}      
\item \textbf{Zero-Shot Prompting: }Provides the model only task instructions, without examples. Prompts for Level 1 relevance classification (Appendix \ref{app:B1}) and Level 2 task classification (Appendix \ref{app:B3}) include inclusion/exclusion criteria and contextual information on SLR stage. This baseline (1) tests the model’s capacity to generalize from task descriptions alone and (2) reflects the most scalable deployment scenario, as real-world pipelines often begin with minimal customization.
\setlength\itemsep{4pt}      
\item \textbf{Few-Shot Prompting: }Supplements task instructions with a small number of annotated examples. This examines whether domain-specific exemplars enhance model calibration, thus simulating cases where limited labelled data are available.
\setlength\itemsep{4pt}      
\item \textbf{Chain-of-Thought (CoT) Prompting: }Instructs the model to generate step-by-step reasoning prior to classification. CoT prompts for Level 1 (Appendix \ref{app:B2}) and Level 2 classifications (Appendix \ref{app:B4}) encourage logical consistency, transparency, and better handling of ambiguous cases, which are qualities essential for literature screening.
\setlength\itemsep{4pt}      
\item \textbf{CoT Few-Shot Prompting: }Combines exemplar-based guidance with explicit reasoning, providing worked examples of both process and output. As the most structured strategy, COT few-shot aims to reduce error propagation while balancing precision and recall in complex tasks.
\setlength\itemsep{4pt}      
\item \textbf{Self-Reflection Prompting: }The model produces an initial response, critiques it, and revises it as needed. By appending a reflection check to a CoT prompt (Appendix \ref{app:B5}), self-reflection promotes calibration and error correction in an attempt to minimize false positives and negatives. Though less established, it reflects an emerging direction in meta-cognitive prompting.
\end{itemize}

We paired all prompts with a system message (Appendix \ref{app:B6}) instructing the model to assume the role of an expert in machine learning and SLR automation. This contextual grounding reduced ambiguity, strengthened task alignment, and improved classification consistency. Collectively, such strategies ranged from minimal intervention (zero-shot) to highly structured reasoning and self-correction (CoT few-shot, self-reflection). This breadth enabled us to evaluate (1) overall accuracy and (2) prompting paradigms' effect on the recall–precision trade-off central to SLR automation.

\subsection{Large Language Models Evaluated}
We evaluated six state-of-the-art LLMs drawn from major providers, including OpenAI, Anthropic, Google DeepMind, DeepSeek, and Meta. Table \ref{tab:llms-overview} summarizes the evaluated models, their providers, their intended use contexts, and their key strengths. Our selection included flagship systems (optimized for reasoning and comprehension) as well as lightweight variants (designed for cost efficiency and scalability). This diversity helped us assess (1) overall accuracy and (2) performance–efficiency trade-offs crucial to large-scale screening pipelines. All six models demonstrated strong capabilities in classification, reasoning, and retrieval-style tasks, making them well-suited for systematic review automation. They also boasted complementary strengths (e.g., advanced reasoning, efficiency, open-source adaptability), thereby facilitating a representative and balanced cross-architectural evaluation of prompting strategies' influence over performance. To ensure reproducibility and consistency, we accessed all models via API endpoints under standardized conditions, with requests routed through the OpenRouter multi-provider gateway.

\begin{table}[!htbp] 
\centering
\caption{Overview of Evaluated LLMs for Literature Screening Automation}
\label{tab:llms-overview}
\footnotesize
\setlength{\tabcolsep}{3pt}        
\renewcommand{\arraystretch}{1.15} 
\begin{tabularx}{\linewidth}{@{}%
  >{\raggedright\arraybackslash}p{0.10\linewidth}%
  >{\raggedright\arraybackslash}p{0.10\linewidth}%
  >{\raggedright\arraybackslash}p{0.15\linewidth}%
  >{\raggedright\arraybackslash}p{0.18\linewidth}%
  >{\raggedright\arraybackslash}X@{}}
\toprule
\textbf{Model} & \textbf{Provider} & \textbf{Scale/Variant} & \textbf{Intended Use} & \textbf{Key Strengths} \\
\midrule
\makecell[tl]{GPT-4o} & OpenAI & Flagship &
Advanced reasoning, comprehension &
Strong performance on reasoning\mbox{-}intensive tasks; widely benchmarked and documented \\
\midrule
\makecell[tl]{GPT-4o-\\mini} & OpenAI & Lightweight &
Cost-efficient, scalable screening &
Reduced computational cost; suitable for high-volume abstract screening \\
\midrule
\makecell[tl]{DeepSeek-\\Chat-v3} & DeepSeek & Mid-scale, efficiency-focused &
General-purpose with efficiency emphasis &
Optimized for efficiency and task alignment; strong balance of cost and performance \\
\midrule
\makecell[tl]{Gemini-\\2.5-Flash} & Google DeepMind & Flagship &
High-performance research applications &
Competitive recall performance; strong comprehension in complex tasks \\
\midrule
\makecell[tl]{Claude-\\3.5-Haiku} & Anthropic & Lightweight &
Fast, cost-efficient text classification &
Compact and efficient; strong reasoning ability relative to size \\
\midrule
\makecell[tl]{Llama-4-\\Maverick} & Meta & Research / opensource &
Reproducible academic and applied research &
Widespread adoption in open-source community; adaptable and transparent \\
\bottomrule
\end{tabularx}
\end{table}

\subsection{Evaluation Metrics}
We evaluated model predictions against our expert-annotated dataset using four widely adopted classification metrics:
\begin{itemize}
\setlength\itemsep{4pt}      
\item \textbf{Accuracy} – the overall proportion of correctly classified records.
\setlength\itemsep{4pt}      
\item \textbf{Precision} – the proportion of true positives among all positive predictions (model specificity).
\setlength\itemsep{4pt}      
\item \textbf{Recall} – the proportion of true positives among all actual positives (model sensitivity).
\setlength\itemsep{4pt}      
\item \textbf{F1-Score} – the harmonic mean of precision and recall, balancing both dimensions.
\end{itemize}

For each LLM–prompting strategy combination, these computed metrics allowed cross-model and cross-prompt comparisons. Overall, our methodology provides a rigorous and reproducible basis for assessing interactions between prompting strategy and LLM architecture in literature-screening automation.

\section{Results}
We organize our results by the two classification levels used to construct our ground truth dataset (Figure \ref{fig:screeningProcess}). The first part reports on \textbf{Level 1 classification}, which determines a study's relevance to SLR automation. The second part covers \textbf{Level 2 classification}, which (1) identifies specific SLR stages addressed (searching, screening, retrieval, synthesis, writing) and (2) detects LLM use. Each part includes subsections on overall performance, comparisons across prompting strategies and models, and analyses of their interactions. Across both levels, we evaluate six LLMs (GPT-4o, GPT-4o-Mini, DeepSeek-Chat-V3, Gemini-2.5-Flash, Claude-3.5-Haiku, and Llama-4-Maverick) under four prompting strategies (zero-shot, few-shot, CoT, and CoT-few-shot). We include self-reflection prompting only for Level 1 classification, given its weaker performance in preliminary testing. We conduct our evaluation using four standard metrics (accuracy, precision, recall, and F1 score), and we supplement it with  descriptive and statistical sub-analyses to capture differential effects of model choice and prompting strategy.

\subsection{Classification of Relevance to SLR Automation}
\begin{figure}[!htbp]
  \centering
  \includegraphics[width=1.0\textwidth]{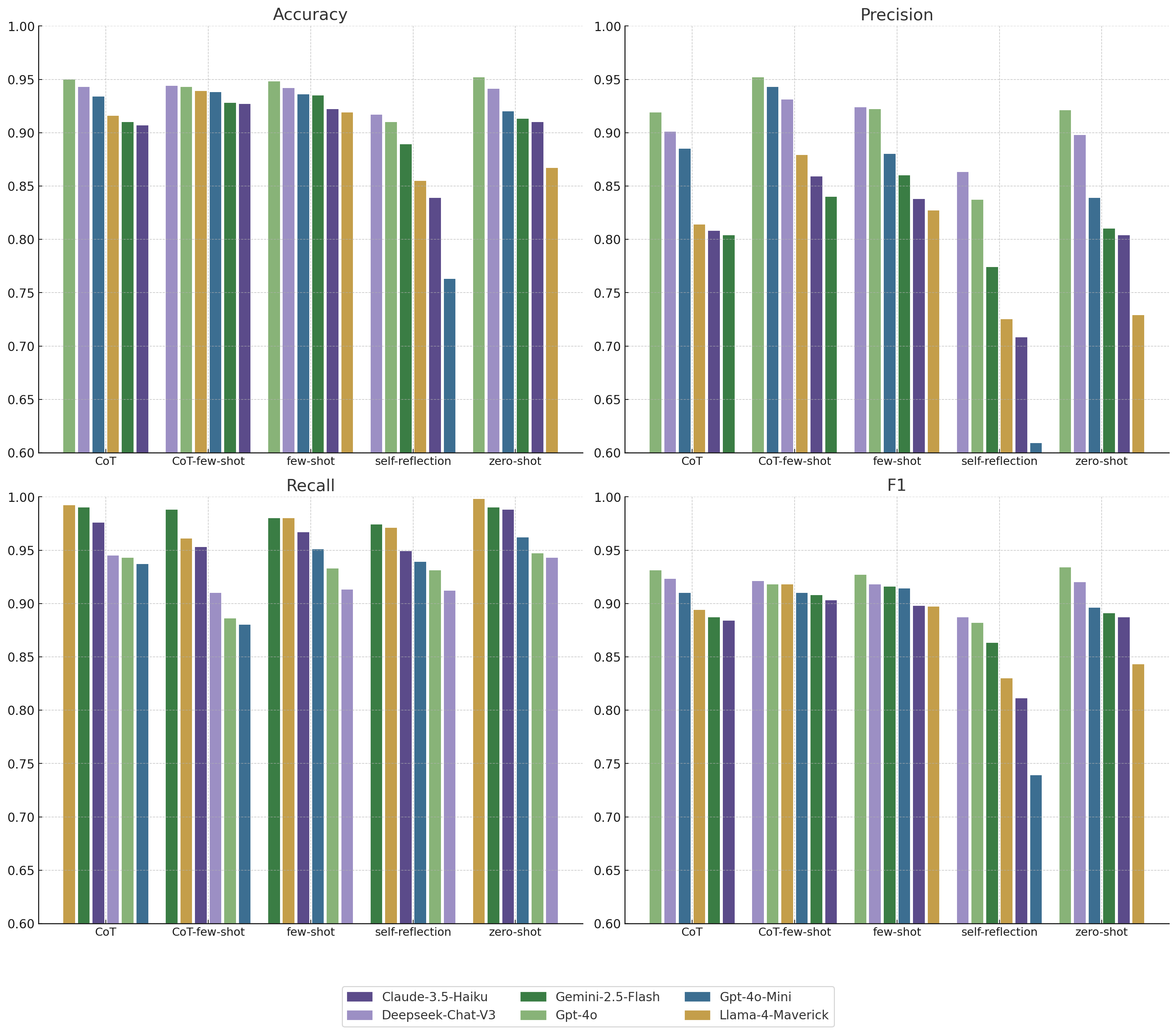}
  \caption{Performance for Prompt-Model Combinations in Relevance Classification}
  \label{fig:promptModelRelev}
\end{figure}

The following evaluations granularly assess how prompt design and model choice jointly shape automated identification of relevant studies (Figure \ref{fig:screeningProcess}, Level 1 Classification). Table \ref{tab:PromtModelCombinaRelevan} (in Appendix A) reports performance metrics for each prompt–model combination, while Figure \ref{fig:promptModelRelev} summarizes these results, enabling systematic comparisons across prompt types and models in terms of accuracy, precision, recall, and F1 score. Appendix \ref{app:B} presents the corresponding prompt templates.

\subsubsection{Effects of Prompting Strategies on Relevance Classification}
Prompting strategies substantially shape LLMs' predictive behaviour in SLR automation. To disentangle prompt design's effects from model-specific characteristics, we first analyse prompt types' performance variations within each LLM before conducting a macro-averaged evaluation across all models. This two-stage approach allows us to assess (1) model-level sensitivity to prompt choice and (2) general trends in prompting effectiveness across the entire LLM ensemble.

\paragraph{Classification Performance Variation of Prompt Types by LLMs}
We performed Friedman’s nonparametric, rank-based test over six datasets to evaluate different prompt types' classification performance variation within each model. Each dataset corresponded to a single LLM and contained predictions from five prompt types, where each observation represented an article and the prediction outcome was coded as true positive (TP), true negative (TN), false positive (FP), or false negative (FN). The results indicate highly significant differences in predictive performance across prompt types for Claude-3.5-Haiku ($\chi^2(4)=134.6,\ \mathrm{p}<1.0\times10^{-25}$), GPT-4o-Mini ($\chi^2(4)=40.7,\ \mathrm{p}<1.0\times10^{-7}$), Gemini-2.5-Flash ($\chi^2(4)=38.3,\ \mathrm{p}<1.0\times10^{-7}$), and Llama-4-Maverick ($\chi^2(4)=27.4,\ \mathrm{p}<1.0\times10^{-5}$). By contrast, prompt-type effects were weaker for GPT-4o ($\chi^2$(4) = 8.2, p = 0.085) and negligible for DeepSeek-Chat-V3 ($\chi^2$(4) = 1.2, p = 0.88). We then calculated Nemenyi mean ranks of prompt types for each model, confirming that the ordering and magnitude of ranks were consistent with the accuracy levels of corresponding prompt types (Table \ref{tab:PromtModelCombinaRelevan}). Across all LLMs, the self-reflection prompt type consistently ranked lowest, whereas the relative ordering of CoT, CoT-few-shot, few-shot, and zero-shot varied across models. These findings suggest that prompt design substantially shapes some LLMs' performance (notably that of Claude-3.5-Haiku, GPT-4o-Mini, Gemini-2.5-Flash, and Llama-4-Maverick), while other LLMs (like GPT-4o and DeepSeek-Chat-V3) exhibit more stable behaviour across different prompt types.

\paragraph{Macro-Averaged Evaluation Metrics of Prompt Types Across Models}
To isolate prompt design's effects from model-specific variation, we examined macro-averaged performance across all models (Table \ref{tab:RelevAveMaticEachPromType}). Prompting strategies (types) exert a substantial influence on classification outcomes.

Combining CoT with few-shot prompting achieved the highest overall performance (F1 = 0.913), followed closely by few-shot (0.912) and CoT (0.905). This confirms that by integrating structured reasoning with exemplars, one can stabilize inferences and mitigate error propagation. There exists a clear recall-precision trade-off. Zero-shot prompting maximized recall (0.971), thereby reducing false negatives (critical in early-stage screening, where excluding relevant studies is most costly); however, precision decreased (0.834), rendering more false positives. By contrast, CoT-few-shot achieved a better balance, raising precision substantially (0.901) while maintaining strong recall (0.930) and producing the best F1. Self-reflection prompting performed worst, with the lowest F1 (0.835) and worst precision (0.753). By encouraging each model to critique its own output, this strategy was designed to improve calibration; nevertheless, in practice, it amplifies over-inclusivity, causing high false-positive rates without compensatory gains in recall.

Overall, prompt design governs the trade-off between inclusivity and efficiency. Zero-shot may be appropriate in high-sensitivity contexts that prioritize recall, but CoT-few-shot offers the most reliable general-purpose strategy by combining robustness with reduced reviewer burden.
\begin{table}[!htbp]
\caption{Macro-Averaged Evaluation Metrics for Each Prompt Type (across all models)}
\label{tab:RelevAveMaticEachPromType}
\centering
\begin{tabular}{lcccc}
\toprule
\textbf{Prompt Type} & \textbf{Accuracy} & \textbf{Precision} & \textbf{Recall} & \textbf{F1} \\
\midrule
CoT              & 0.927 & 0.855 & 0.964 & 0.905 \\
CoT-few-shot     & \textbf{0.937} & \textbf{0.901} & 0.930 & \textbf{0.913} \\
Few-shot         & 0.934 & 0.875 & 0.954 & 0.912 \\
Self-reflection  & 0.862 & 0.753 & 0.946 & 0.835 \\
Zero-shot        & 0.917 & 0.834 & \textbf{0.971} & 0.895 \\
\bottomrule
\end{tabular}
\end{table}

\subsubsection{Effects of Model Choice on Relevance Classification}
While prompt design can indeed drive predictive outcomes, choice of LLM can also introduce systematic variation. To disentangle the effects of model architecture and training from those of prompting strategies, we evaluate each LLM's performance under identical prompt conditions. This section analyses model-level variation for each respective prompt type before aggregating results to examine macro-level patterns across all prompting strategies.

\paragraph{Classification Performance Variation of LLMs by Prompt Types}
To assess whether classification performance varies systematically across LLMs under different prompting strategies, we applied Friedman’s test to five datasets, each corresponding to a specific prompt type. In this design, each dataset contained predictions from six LLMs, with each observation coded as TP, TN, FP, or FN. The results reveal pronounced differences in model performance under most prompting conditions. The self-reflection setting produced the largest effect ($\chi^2(5)=786.8,\ \mathrm{p}=8.1\times10^{-168}$
), highlighting exceptionally strong cross-model variation. We also observed substantial heterogeneity under CoT prompting ($\chi^2(5)=177.7,\ \mathrm{p}=1.7\times10^{-36}$
) and zero-shot prompting ($\chi^2(5)=120.3,\ \mathrm{p}=2.8\times10^{-24}$
). Few-shot prompting exhibited a more moderate, though still very significant, effect ($\chi^2(5)=24.1,\ \mathrm{p}=2.\times10^{-4}$
). By contrast, CoT-few-shot yielded only marginal evidence of cross-model variation ($\chi^2(5)=11.5,\ \mathrm{p}=0.042$).

For each prompt type, we then computed each LLM's Nemenyi mean rank; the ranks' ordering and magnitude aligned with observed accuracy levels (Table \ref{tab:PromtModelCombinaRelevan}). Furthermore, according to the Nemenyi ranks, self-reflection and zero-shot prompts generated the clearest cross-LLM stratification; frontier models (like GPT-4 variants) consistently achieved top ranks while smaller models trailed behind. Under CoT and few-shot prompting, rank differences were evident but less pronounced; meanwhile, CoT-few-shot minimized discrepancies, producing nearly indistinguishable model rankings. These findings indicate that model choice matters most for self-reflection and CoT prompting, while cross-model variation becomes less pronounced under CoT-few-shot and few-shot conditions.

\paragraph{Macro-Averaged Evaluation Metrics of LLMs across Prompt Types}
To isolate choice-of-model effects from prompt-specific variations, we examined macro-averaged performance across all prompt types. In doing so, we took a broader look at the way model selection influences classification outcome when averaged across prompting strategies. Indeed, our analysis showcased (1) consistently strong performers and (2) models whose performances depended heavily on prompts.

As shown in Table \ref{tab:RevelaAveMetricEachModel}, GPT-4o achieved the highest macro-F1 (0.918), closely followed by DeepSeek-Chat-V3 (0.914). Gemini-2.5, Claude-3.5-Haiku, Llama-4-Maverick, and GPT-4o-mini formed a second tier, with macro-F1 scores between 0.874 and 0.893. Model profiles reveal systematic trade-offs. Gemini-2.5, Llama-4-Maverick, and Claude-3.5-Haiku attained exceptionally high recall (0.967–0.984) but markedly lower precision (0.795–0.818), underscoring the risk of inflated false positives. These models therefore rely heavily on precision-enhancing prompts (e.g., CoT-few-shot) to achieve balanced performance. By contrast, GPT-4o and DeepSeek delivered more stable results, ranking highest in accuracy, precision, and F1 while maintaining strong recall (>0.925). Visual patterns in Figure \ref{fig:promptModelRelev} confirm these trends: GPT-4o and DeepSeek consistently occupied the upper band for accuracy, precision, and F1, whereas Gemini and Llama dominated in recall but underperformed in precision. Hence, the “best” model varies with evaluation priority: GPT-4o and DeepSeek provide the most reliable all-round performance, while Gemini and Llama may best suit those who (1) prioritize inclusivity and (2) manage false positives with stricter prompting.

\begin{table}[!htbp]
\caption{Macro-Averaged Evaluation Metrics For Each Model (across all prompt types)}
\label{tab:RevelaAveMetricEachModel}
\centering
\begin{tabular}{lcccc}
\toprule
\textbf{Model Type} & \textbf{Accuracy} & \textbf{Precision} & \textbf{Recall} & \textbf{F1} \\
\midrule
Claude-3.5-Haiku   & 0.901 & 0.803 & 0.967 & 0.877 \\
Deepseek-Chat-V3   & 0.938 & 0.904 & 0.925 & 0.914 \\
Gemini-2.5         & 0.915 & 0.818 & \textbf{0.984} & 0.893 \\
Gpt-4o             & \textbf{0.941} & \textbf{0.910} & 0.928 & \textbf{0.918} \\
Gpt-4o-Mini        & 0.898 & 0.831 & 0.934 & 0.874 \\
Llama-4-Maverick   & 0.899 & 0.795 & 0.980 & 0.876 \\
\bottomrule
\end{tabular}
\end{table}

\subsubsection{Interactions Between Prompting and Models}
\begin{figure}[!htbp]
  \centering
  \includegraphics[width=0.8\textwidth]{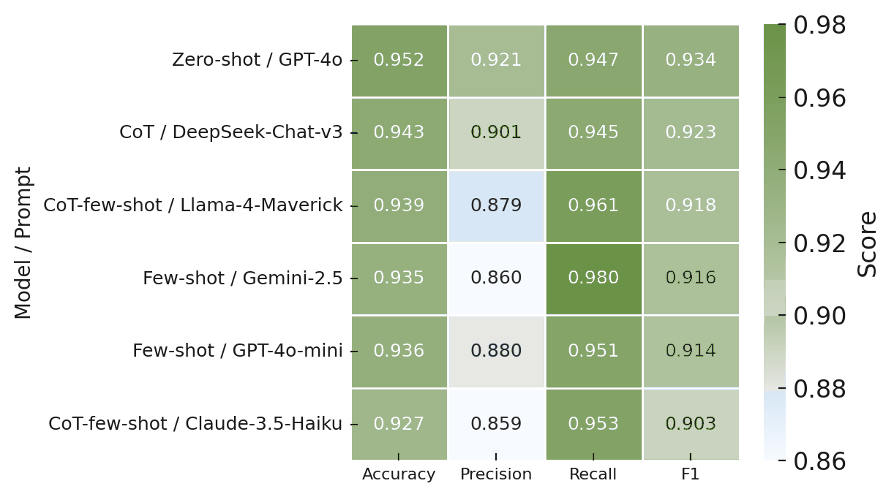}
  \caption{Evaluation Metrics Of Best Performing Prompt Types By Model (relevance classification)}
  \label{fig:bestPromptModelRelev}
\end{figure}

According to our evaluation of top-performing prompt–model combinations (Table \ref{tab:BestPromptByModel}; Figures \ref{fig:bestPromptModelRelev}), optimal prompting strategy varies by model. With zero-shot prompting, GPT-4o achieved the most balanced performance, recording the highest accuracy (0.952), precision (0.921), and F1 (0.934). With few-shot prompting, Gemini-2.5 attained the strongest recall (0.980), identifying nearly all relevant studies; however, it did so at the cost of lower precision (0.860) (i.e., more false positives). Similarly, GPT-4o-mini and Gemini-2.5 both benefited from exemplar guidance paired with few-shot prompting. With CoT and CoT-few-shot respectively, DeepSeek-Chat-v3 and Llama-4-Maverick also delivered strong recall (0.945 and 0.961, respectively); however, they experienced modest reductions in precision (0.901 and 0.879, respectively). These results suggest that DeepSeek aligns well with structured reasoning, while Llama-4-Maverick requires both reasoning and exemplar guidance to offset its recall bias. Claude-3.5-Haiku also favoured CoT-few-shot, but lagged behind in precision (0.859) and F1 (0.903).

No single prompting strategy dominated across models. While CoT-few-shot provided the most reliable baseline, F1 saw measurable gains (0.01–0.02) when prompts were tailored to model-specific characteristics, and this adjustment could scale substantially to large screening pipelines. Across configurations, recall consistently exceeded precision; as such, successful LLM–prompt pairing may improve identification of relevant literature but may not improve exclusion of irrelevant literature. Strategic alignment remains important, too: with zero-shot, GPT-4o offered the most balanced accuracy; with few-shot, Gemini-2.5 maximized recall. Meanwhile, other pairings reveal trade-offs that researchers may leverage to minimize false negatives or false positives.

\subsubsection{Confidence Intervals of Evaluation Metrics}
As shown in Table \ref{tab:PromtModelCombinaRelevan}, we report 95\% confidence intervals (CIs) for all metrics. We estimated Accuracy, Precision, and Recall CIs with Wilson score intervals, and we obtained F1 CIs with bootstrap resampling at the abstract level (5,000 replicates) to account for the joint variability of Precision and Recall. A large denominator made accuracy intervals narrow, whereas Precision and Recall spanned wider ranges, reflecting class imbalance. F1 intervals were typically broader than Accuracy intervals but provided realistic measures of uncertainty. These results indicate that small differences (<1\%) between prompts and models often fall within overlapping CIs, underscoring the need for cautious interpretation and complementary paired tests.

\subsection{Classification of SLR Stage(s) and LLM Use}
The following evaluations assess how variations in prompt design and model selection influence performance across six classification tasks (Figure \ref{fig:screeningProcess}): five SLR automation stages (searching, screening, retrieval, synthesis, and writing) and detection of LLM use. We report results as (1) task-specific model and prompt sensitivities; and (2) macro-averages across conditions (where appropriate). Given our dataset composition (491 relevant articles with notable task-specific imbalances, i.e., 260 articles about screening and only 34 about writing), results interpretation must account for task prevalence.

\subsubsection{Effects of Prompting Methods on Tasks Classification Outcomes}
Prompt design and model choice jointly shape performance across six Level-2 tasks (searching, screening, retrieval, synthesis, writing, and detecting LLM use). To disentangle these effects, we first assess task-specific prompt sensitivity within each model using Friedman’s nonparametric test. In doing so, we identify points across tasks where prompt design significantly alters performance. We then report prompt types' macro-averaged performance metrics across models and tasks, highlighting systematic precision–recall trade-offs for each prompting strategy.

\paragraph{Classification Performance of Prompt Types by Models across Tasks}
For each classification task and within each LLM, we applied Friedman’s nonparametric, rank-based test to determine whether prompt type significantly influenced performance. Each dataset corresponded to a single model and contained predictions from four prompt types, coded as TP, TN, FP, or FN. As summarized in Table \ref{tab:task-llm-friedman}, our results reveal substantial variation across tasks and models. In retrieval, some models were sensitive to prompting but  others were largely unaffected. Screening exhibited broader prompt effects, with several models responding strongly to different prompt types. Searching was prompt-sensitive in select models but not in others. Prompting also affected synthesis, but only in a subset of models. Writing was most sensitive to prompting, with multiple LLMs showing significant differences across prompt types. Finally, in the “Using LLMs” classification task, prompt effects were evident but limited to specific models. In sum, our findings suggest that prompt sensitivity is unevenly distributed across tasks and models. Claude-3.5-Haiku and GPT-4o-Mini consistently displayed strong responsiveness to prompt design, Gemini-2.5-Flash and GPT-4o were largely prompt-invariant, and DeepSeek-Chat-V3 and Llama-4-Maverick exhibited task-dependent variation.

For each classification task and each model, we computed Nemenyi mean ranks of prompt types; the ordering of ranks aligned with observed accuracy levels (\ref{tab:tasks-prompts-models-ci}). Across the six SLR automation tasks, mean ranks revealed systematic yet task-dependent variation in prompt types' relative effectiveness. CoT and CoT-few-shot generally achieved more favourable ranks, especially in reasoning-intensive tasks like retrieval and synthesis. Zero-shot and few-shot often ranked lower, though their performance varied by task and model. LLM choice also interacted with task type: frontier models (like GPT-4 variants and Gemini-2.5-Flash) maximized advantages of structured prompting, whereas smaller models (like Claude-3.5-Haiku and Llama-4-Maverick) were less differentiated. Task demands further shaped rankings: reasoning-heavy tasks amplified the benefits of structured prompting, while simpler screening tasks minimized differences. These findings indicate that both model architecture and task characteristics jointly determine prompt types' relative success; as such, prompting strategies should be tailored to specific SLR automation contexts.

\definecolor{sig}{HTML}{E7F4EA} 

\begin{table}[!htbp]
\centering
\setlength{\tabcolsep}{4pt}
\renewcommand{\arraystretch}{1.08}
\scriptsize
\caption{Friedman Statistics ($\chi^2$ with df = 3) and $p$–values Across Tasks and LLMs}
\label{tab:task-llm-friedman}
\begin{adjustbox}{width=\linewidth} 
\begin{tabular}{l *{6}{cc}}
\toprule
 & \multicolumn{2}{c}{Claude-3.5-Haiku}
 & \multicolumn{2}{c}{DeepSeek-Chat-V3}
 & \multicolumn{2}{c}{Gemini-2.5-Flash}
 & \multicolumn{2}{c}{GPT-4o-Mini}
 & \multicolumn{2}{c}{GPT-4o}
 & \multicolumn{2}{c}{Llama-4-Maverick} \\
\cmidrule(lr){2-3}\cmidrule(lr){4-5}\cmidrule(lr){6-7}\cmidrule(lr){8-9}\cmidrule(lr){10-11}\cmidrule(lr){12-13}
\textbf{Task} & $\chi^2$ & $p$ & $\chi^2$ & $p$ & $\chi^2$ & $p$ & $\chi^2$ & $p$ & $\chi^2$ & $p$ & $\chi^2$ & $p$ \\
\midrule
Retrieval
  & \cellcolor{sig}17.18 & \textbf{0.00065}
  & 7.65 & 0.054
  & 1.24 & 0.743
  & \cellcolor{sig}10.31 & \textbf{0.016}
  & 1.63 & 0.653
  & 6.06 & 0.109 \\
Screening
  & \cellcolor{sig}11.97 & \textbf{0.0075}
  & \cellcolor{sig}10.58 & \textbf{0.0143}
  & 4.15 & 0.246
  & \cellcolor{sig}10.50 & \textbf{0.0148}
  & 5.55 & 0.136
  & 6.94 & 0.074 \\
Searching
  & \cellcolor{sig}12.97 & \textbf{0.0047}
  & 0.43 & 0.935
  & 1.78 & 0.619
  & \cellcolor{sig}11.15 & \textbf{0.011}
  & \cellcolor{sig}11.77 & \textbf{0.008}
  & 7.01 & 0.072 \\
Synthesis
  & \cellcolor{sig}9.89  & \textbf{0.0195}
  & 0.33 & 0.954
  & \cellcolor{sig}8.38 & \textbf{0.0387}
  & \cellcolor{sig}16.22 & \textbf{0.0010}
  & 4.32 & 0.229
  & 3.74 & 0.291 \\
Writing
  & \cellcolor{sig}23.77 & \textbf{<\thinspace 0.0001}
  & 6.82 & 0.078
  & 4.10 & 0.250
  & \cellcolor{sig}13.71 & \textbf{0.0033}
  & 2.44 & 0.487
  & \cellcolor{sig}8.59 & \textbf{0.035} \\
Using LLMs
  & \cellcolor{sig}13.22 & \textbf{0.0042}
  & 3.61 & 0.307
  & 0.63 & 0.890
  & \cellcolor{sig}18.63 & \textbf{0.0003}
  & 4.92 & 0.178
  & 7.71 & 0.052 \\
\bottomrule
\end{tabular}
\end{adjustbox}

\vspace{3pt}
\footnotesize\textit{Note.} Friedman Tests compare performance across four prompt types within each LLM for a given task; df $=3$. 
Bold indicates $p<0.05$; highlighted $\chi^2$ matches those significant $p$–values.
\end{table}

\begin{figure}[ht!]
  \centering
  \includegraphics[width=0.7\textwidth]{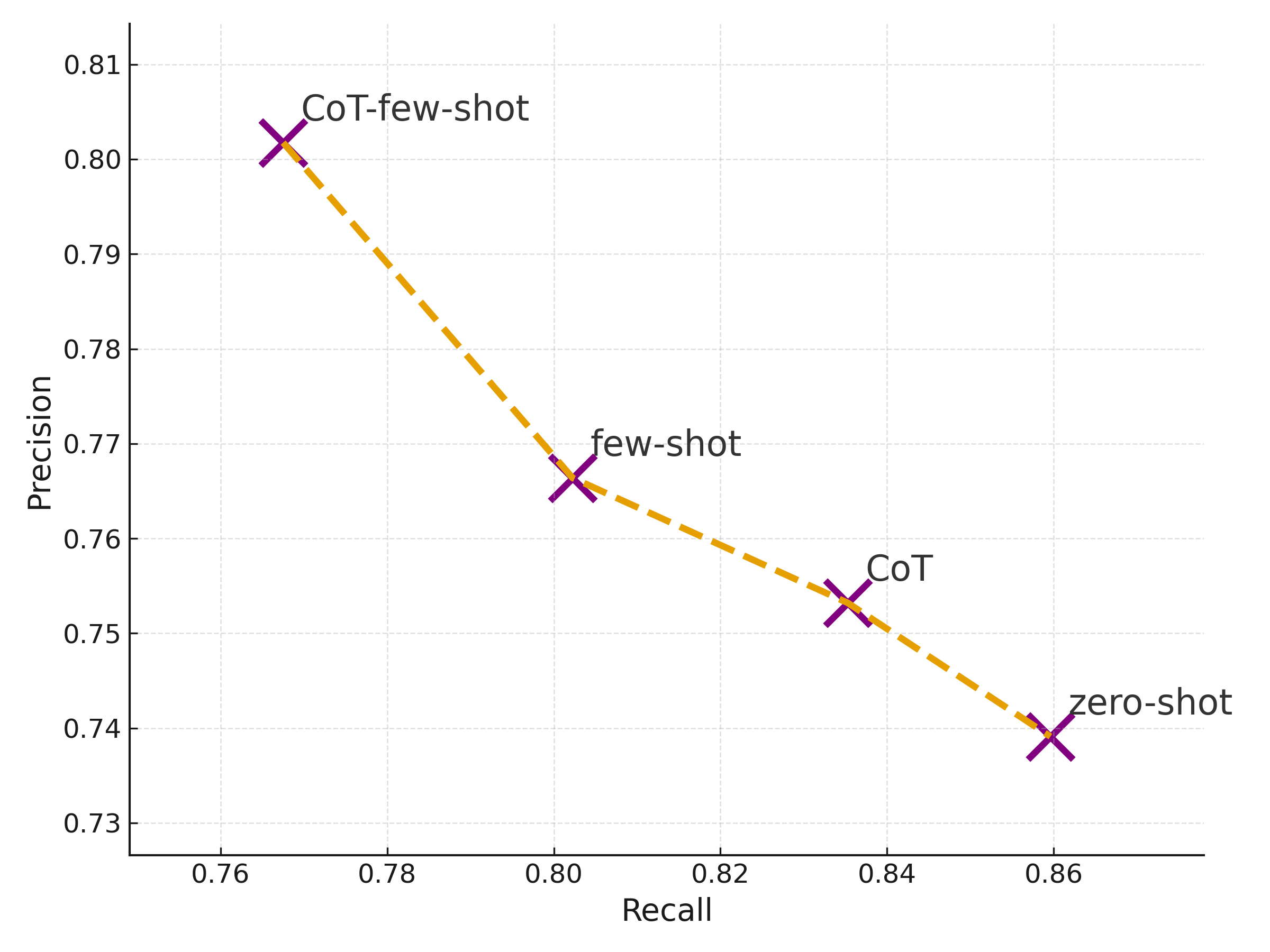}
  \caption{Precision–Recall trade-off for prompt strategies}
  \label{fig:taskPromPreciRecallTradeOff}
\end{figure}

\paragraph{Macro-Averaged Evaluation Metrics of Prompt Types across Models and Tasks}
According to our macro-averaged evaluation across all models and tasks (Table \ref{tab:TaskAverMatricEachPrompt}, Figure\ref{fig:taskPromPreciRecallTradeOff}), prompting strategies systematically shape the recall–precision trade-off. Accuracy remained consistent across methods (0.872–0.877), indicating stable baseline reliability, but precision, recall, and F1 diverged. Zero-shot prompting achieved the highest recall (0.838) and would prove particularly valuable in early-stage screening, where under-inclusion (of relevant literature) would cost more than over-inclusion. However, zero-shot's low precision (0.713) increased false positives and would burden downstream reviewers. In contrast, CoT-few-shot prompting yielded the highest precision (0.774) and a competitive F1 (0.752) by balancing exemplar guidance and reasoning. This gain in precision reduced spurious classifications, though at the cost of moderately lower recall (0.736). Few-shot prompting delivered mid-range performance (F1 = 0.754), especially benefiting models that align well with contextual exemplars. CoT prompting raised recall (0.812) relative to few-shot prompting but sacrificed precision (0.718); it tended toward over-inclusion when no exemplars grounded its reasoning. In short, F1 scores were tightly clustered (0.752–0.766), indicating no uniformly dominant strategy.

The plot in Figure \ref{fig:taskPromPreciRecallTradeOff} reveals a clear precision–recall trade-off across prompt strategies: recall rises toward zero-shot, precision peaks with CoT-few-shot, and few-shot and CoT occupy the middle ground. The dashed line highlights a consistent downward trend from higher-precision/lower-recall (CoT-few-shot) to lower-precision/higher-recall (zero-shot): incremental precision losses typically accompany gains in recall. Practically, strategy should vary with stage: use zero-shot for early, high-recall passes; transition to few-shot or CoT to balance workload and errors; and deploy CoT-few-shot when precision is paramount. This progression aligns model behaviour with screening objectives, reducing overall workload while controlling risk of error.

\begin{table}[!htbp]
\centering
\caption{Macro-Averaged Evaluation Metrics for Each Prompt (across models \& tasks)}
\label{tab:TaskAverMatricEachPrompt}
\begin{tabular}{lcccc}
\toprule
\textbf{Prompt Type} & \textbf{Accuracy} & \textbf{Precision} & \textbf{Recall} & \textbf{F1} \\
\midrule
CoT          & 0.872 & 0.718 & 0.812 & 0.758 \\
CoT-few-shot & \textbf{0.877} & \textbf{0.774} & 0.736 & 0.752 \\
Few-shot     & \textbf{0.877} & 0.741 & 0.775 & 0.754 \\
Zero-shot    & 0.876 & 0.713 & \textbf{0.838} & \textbf{0.766} \\
\bottomrule
\end{tabular}
\end{table}

\subsubsection{Effect of LLM Choice on Tasks Classification Outcomes}
Beyond assessing individual models' prompt sensitivities, each LLM's performance need also be evaluated under the same prompting conditions. As such, we apply Friedman’s nonparametric test to measure cross-model variation for each task, with datasets organized by prompt type. A macro-level LLM evaluation complements our analysis by revealing (1) precision–recall trade-offs and (2) shifts in overall model effectiveness when averaged across prompt types and tasks.

\paragraph{Classification Performance of LLMs by Prompt Types across Tasks}
To evaluate systematic differences in LLM performance across prompting strategies, we applied Friedman’s nonparametric test to four datasets for each classification task; each dataset corresponded to a distinct prompt type and contained predictions from six models. Once again, we coded each outcome as TP, TN, FP, or FN.

As shown in Table \ref{tab:task-prompt-friedman}, Friedman tests across six SLR automation tasks suggest that cross-LLM variation depends strongly on prompt type. Structured prompts consistently amplified model differences, whereas minimal prompts often produced more uniform outcomes. For retrieval, searching, synthesis, and writing, we observed significant cross-model variation under CoT and CoT-few-shot prompting, while few-shot and zero-shot prompting rendered no significant differences. In contrast, screening revealed strong LLM variability across all prompt types (all $p$ < 0.001); thus, model choice remains critical to screening, with or without regard to prompt design. The “Using LLMs” task also exposed broad sensitivity; models exhibited significant differences under all prompting conditions (including minimal prompting). Taken together, these results indicate that LLM performance depends heavily on prompting. Practically, (1) CoT and CoT-few-shot expose greater performance gaps across most tasks, while (2) few-shot and zero-shot generally reduce cross-model variation, but not for tasks inherently more sensitive to model design (screening, detection of LLM use).

For each classification task, we computed Nemenyi mean ranks of LLMs; our rank ordering reflected corresponding accuracy levels (Table \ref{tab:tasks-prompts-models-ci}{}). Across the six SLR automation tasks and across prompt types, model performance exhibited consistent yet task-dependent variation. GPT-4o and Gemini-2.5-Flash frequently achieved the best ranks, particularly in reasoning-oriented tasks (retrieval, synthesis classification), while smaller models like Claude-3.5-Haiku and Llama-4-Maverick generally placed lower (though their relative positions shifted by task and prompt). Prompt type mattered most for complex tasks, where structured strategies (CoT, CoT-few-shot) accentuated performance gaps and simpler strategies (zero-shot, few-shot) narrowed differences. These results suggest that model design and task complexity both interact with prompting strategy to shape comparative performance; thus, it remains important to benchmark LLMs across diverse task settings.

\definecolor{sig}{HTML}{E7F4EA} 

\begin{table}[!htbp]
\centering
\setlength{\tabcolsep}{4pt}
\renewcommand{\arraystretch}{1.08}
\scriptsize
\caption{Friedman Statistics ($\chi^2$ with df = 5) and $p$–values Across Tasks and Prompt Types}
\label{tab:task-prompt-friedman}
\begin{adjustbox}{width=\linewidth} 
\begin{tabular}{l *{4}{c l}}
\toprule
 & \multicolumn{2}{c}{CoT}
 & \multicolumn{2}{c}{CoT-few-shot}
 & \multicolumn{2}{c}{Few-shot}
 & \multicolumn{2}{c}{Zero-shot} \\
\cmidrule(lr){2-3}\cmidrule(lr){4-5}\cmidrule(lr){6-7}\cmidrule(lr){8-9}
\textbf{Task} & $\chi^2$ & $p$ & $\chi^2$ & $p$ & $\chi^2$ & $p$ & $\chi^2$ & $p$ \\
\midrule
Retrieval
  & \cellcolor{sig}28.78 & \textbf{<0.0001}
  & \cellcolor{sig}26.88 & \textbf{<0.0001}
  & 1.84 & 0.871
  & 9.10 & 0.105 \\
Screening
  & \cellcolor{sig}31.15 & \textbf{<0.00001}
  & \cellcolor{sig}29.35 & \textbf{<0.00002}
  & \cellcolor{sig}23.42 & \textbf{<0.001}
  & \cellcolor{sig}25.13 & \textbf{<0.001} \\
Searching
  & \cellcolor{sig}45.74 & \textbf{<0.00001}
  & \cellcolor{sig}22.36 & \textbf{<0.001}
  & 7.77 & 0.169
  & 5.69 & 0.337 \\
Synthesis
  & \cellcolor{sig}14.53 & \textbf{0.0126}
  & \cellcolor{sig}26.14 & \textbf{<0.001}
  & 6.70 & 0.244
  & 4.89 & 0.429 \\
Writing
  & \cellcolor{sig}40.88 & \textbf{<0.00001}
  & \cellcolor{sig}26.37 & \textbf{<0.001}
  & 5.89 & 0.317
  & 4.11 & 0.533 \\
Using LLMs
  & \cellcolor{sig}14.30 & \textbf{0.0138}
  & \cellcolor{sig}40.09 & \textbf{<0.0001}
  & \cellcolor{sig}31.84 & \textbf{<0.0001}
  & \cellcolor{sig}43.61 & \textbf{<0.0001} \\
\bottomrule
\end{tabular}
\end{adjustbox}

\vspace{3pt}
\footnotesize\textit{Note.} Bold indicates $p<0.05$; highlighted $\chi^2$ corresponds to significant $p$–values.
\end{table}

\paragraph{Macro Performance and Precision–Recall Trade-offs by Model}

\begin{figure}[!htbp]
  \centering
  \begin{subfigure}[t]{0.49\textwidth}
    \centering
    \includegraphics[height=5.8cm, keepaspectratio]{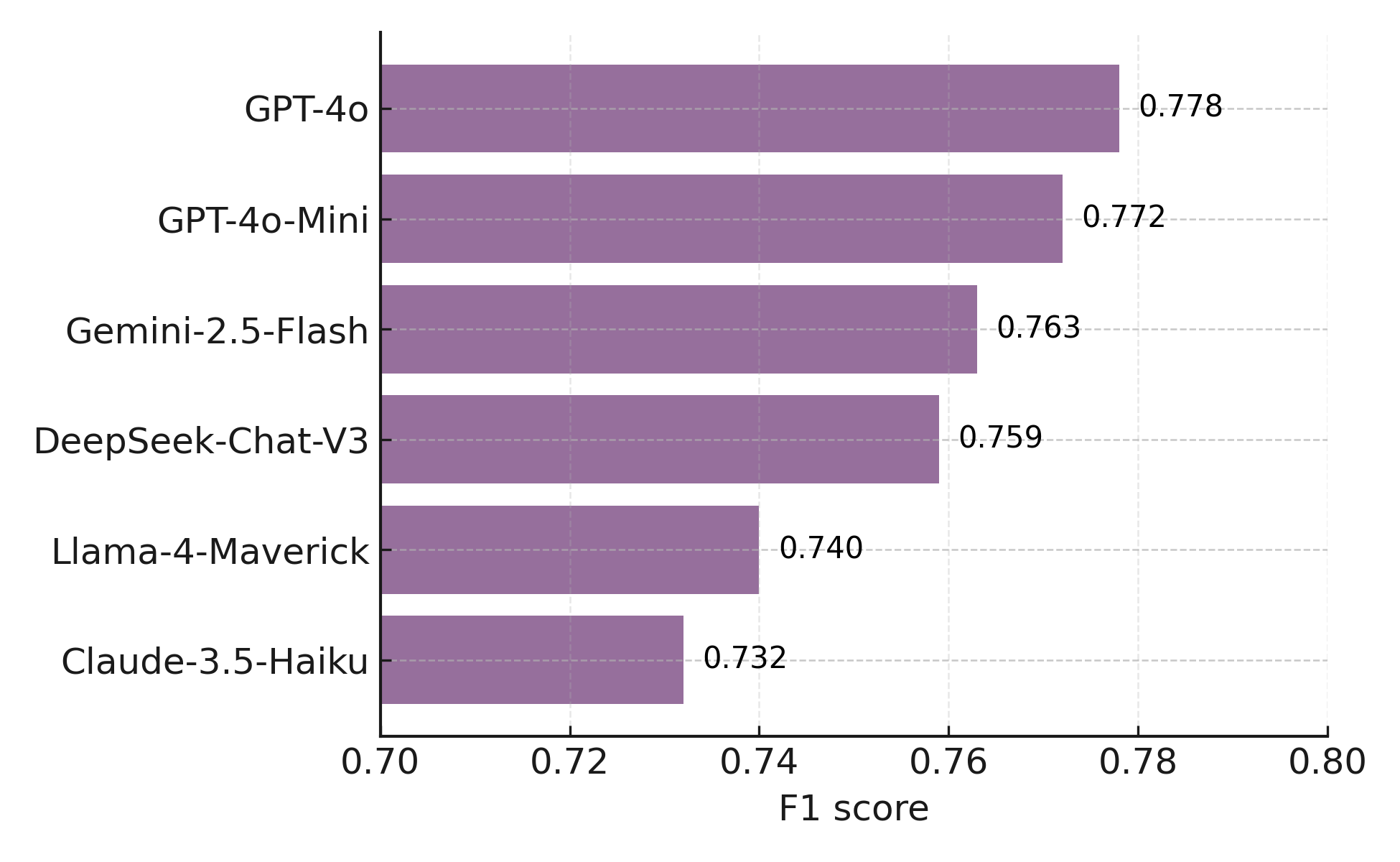}
    \caption{F1 by model}\label{fig:f1-bars}
  \end{subfigure}
  \hfill
  \begin{subfigure}[t]{0.49\textwidth}
    \centering
    \includegraphics[height=5.8cm, keepaspectratio]{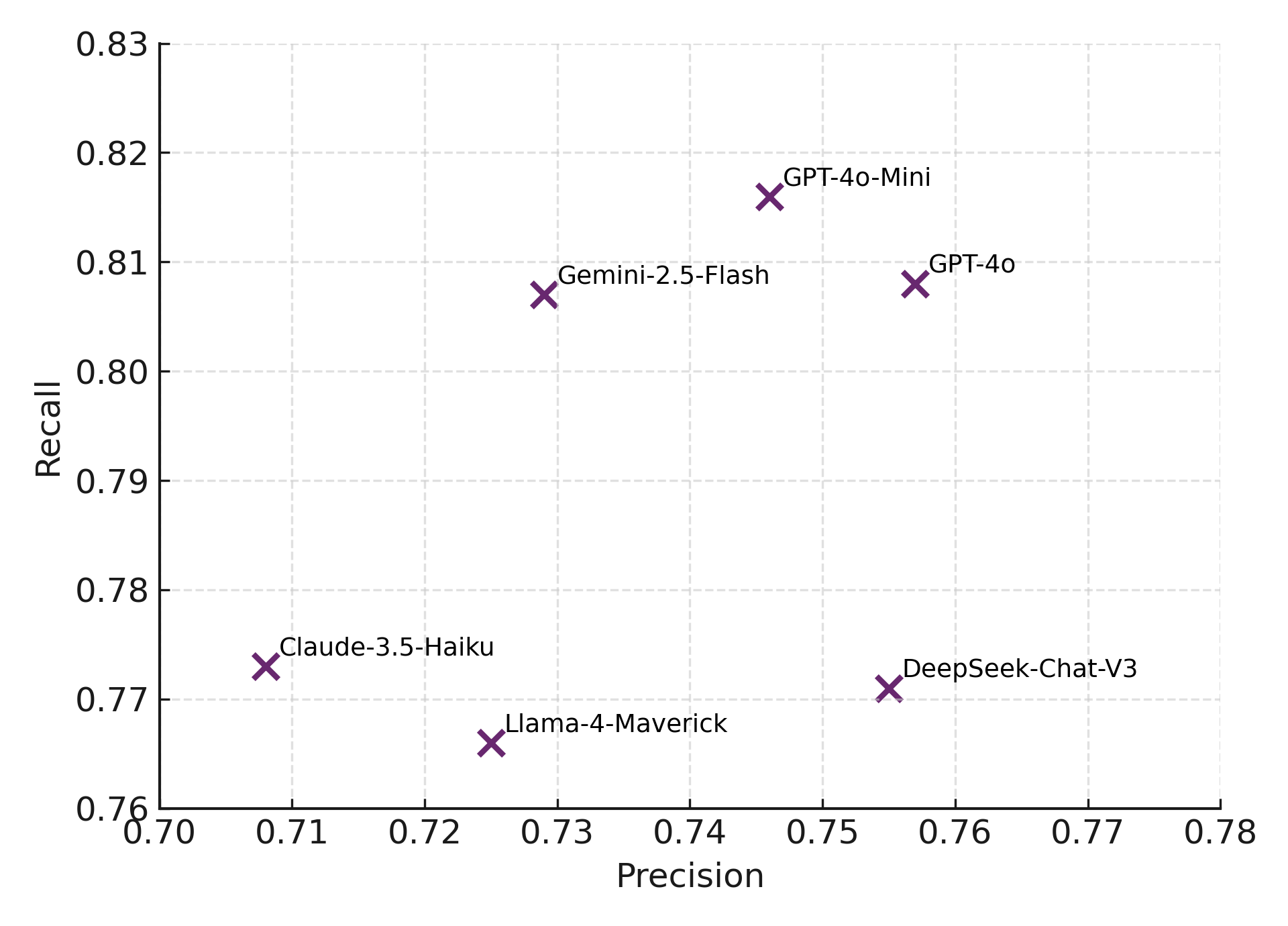}
    \caption{Precision–Recall by model}\label{fig:pr-scatter}
  \end{subfigure}
  \caption{F1 \& Precision–Recall Trade-off Across Models}\label{fig:side-by-side}
\end{figure}

Model selection significantly influences classification outcome, particularly by shaping the precision–recall trade-off (Table \ref{tab:TaskAveraMetriEachModel}, Figure \ref{fig:f1-bars}, and Figure\ref{fig:pr-scatter}). GPT-4o delivered the strongest overall performance (accuracy = 0.886, F1 = 0.778), combining the highest precision (0.757) with strong recall (0.808). Its compact variant, GPT-4o-mini, achieved comparable balance (F1 = 0.772). Gemini-2.5-Flash (F1 = 0.763) and DeepSeek-Chat-v3 (F1 = 0.759) constituted a second performance tier. Gemini emphasized recall (0.807) at the cost of lower precision (0.729), while DeepSeek provided more balance (precision = 0.755, recall = 0.771). Llama-4-Maverick (F1 = 0.740) and Claude-3.5-Haiku (F1 = 0.732) consistently underperformed across prompt types, with weaker precision and greater instability. In all models, accuracy and recall exceeded precision, which suggests a systematic bias toward over-inclusion (i.e., more reliably capturing relevant items than excluding false positives). Overall, OpenAI models (GPT-4o, GPT-4o-mini) stay most stable across tasks, while Gemini and DeepSeek remain competitive but more prompt-sensitive. Claude and Llama are recall-oriented but often sacrifice precision.

The precision–recall trade-off chart (Figure \ref{fig:pr-scatter}) reinforces these patterns. GPT-4o and GPT-4o-mini sustain a strong equilibrium between inclusivity and selectivity. Gemini achieves comparable recall but lower precision, reflecting over-inclusivity, while DeepSeek maintains a mid-range balance. Llama and Claude demonstrate the weakest trade-off, with both precision and recall falling at the lower end. Taken together, no model dominates across dimensions, but GPT-4o and GPT-4o-mini consistently provide the most reliable equilibrium.

\begin{table}[!htbp]
\centering
\caption{Macro-Averaged Evaluation Metrics for Each Model (across prompts \& tasks)}
\label{tab:TaskAveraMetriEachModel}
\label{tab:macro_metrics_models}
\begin{tabular}{lcccc}
\toprule
\textbf{Model} & \textbf{Accuracy} & \textbf{Precision} & \textbf{Recall} & \textbf{F1} \\
\midrule
GPT-4o             & \textbf{0.886} & \textbf{0.757} & 0.808 &\textbf{0.778} \\
GPT-4o-mini        & 0.880 & 0.746 & \textbf{0.816} & 0.772 \\
Gemini-2.5-Flash   & 0.879 & 0.729 & 0.807 & 0.763 \\
DeepSeek-Chat-v3   & 0.878 & 0.755 & 0.771 & 0.759 \\
Llama-4-Maverick   & 0.869 & 0.725 & 0.766 & 0.740 \\
Claude-3.5-Haiku   & 0.860 & 0.708 & 0.773 & 0.732 \\
\bottomrule
\end{tabular}
\end{table}

\subsubsection{Effect of Classification Tasks}
Performance varied markedly across the six classification tasks, reflecting both dataset balance and task complexity (Table \ref{tab:TaskAveragedMatricsEachTask}).

\textbf{High-performing tasks.} Among traditional SLR stages, screening produced the strongest results (F1 = 0.868), consistent with its prevalence in our dataset (260 examples); this provided stable training signals. Retrieval also performed well (F1 = 0.810), balancing precision and recall effectively.

\textbf{Moderate performance.} Searching produced less reliable results (F1 = 0.766), despite a reasonable sample size (147). This is likely because variable semantic descriptions (of “searching") hindered generalization.

\textbf{Low-performing tasks.} Synthesis (F1 = 0.642) and writing (F1 = 0.702) yielded the weakest results. Both were underrepresented in our dataset (109 and 34 examples, respectively) and demanded higher-order reasoning (summarization, integration, generative framing) that was less lexically anchored. Writing also derived unusually high accuracy (0.953) but very low precision (0.640); this suggests systematic over-classification of irrelevant literature.

\textbf{LLM use as a distinct dimension.} LLM-use detection achieved the highest overall performance (F1 = 0.925; accuracy = 0.960). This reflects (1) the ubiquity of such research (127 examples) and (2) the salience of lexical cuing (e.g., “ChatGPT”, “large language model”, “transformer”) to reduce ambiguity. Precision (0.908) and recall (0.946) were jointly high: models seldom missed or over-predicted LLM-related literature.

Generally, task performance follows a gradient from concrete and lexically anchored (LLM use, screening, retrieval) to abstract and conceptually demanding (synthesis, writing). LLM-use detection is most reliably automatable, whereas synthesis and writing remain constrained by data scarcity and semantic abstraction.

\begin{table}[!htbp]
\centering
\caption{Macro-Averaged Evaluation Metrics for Each Task (across prompts \& models)}
\label{tab:TaskAveragedMatricsEachTask}
\begin{tabular}{lcccc}
\toprule
\textbf{Task} & \textbf{Accuracy} & \textbf{Precision} & \textbf{Recall} & \textbf{F1} \\
\midrule
Retrieval    & 0.851 & 0.792 & 0.831 & 0.810 \\
Screening      & 0.861 & \textbf{0.874} & \textbf{0.864} & \textbf{0.868} \\
Searching      & 0.875 & 0.745 & 0.797 & 0.766 \\
Synthesis   & 0.837 & 0.631 & 0.662 & 0.642 \\
Writing        & \textbf{0.953} & 0.640 & 0.797 & 0.702 \\
Use of LLMs    & \textbf{0.960} & \textbf{0.908} & \textbf{0.946} & \textbf{0.925} \\
\bottomrule
\end{tabular}
\end{table}

\subsubsection{Interactions Among Prompts, Models, and Classification Tasks}
While macro-averages isolate the respective effects of prompting strategy and model choice, their interaction shapes overall performance. Each model responds differently to a given prompting method, and each task responds differently to a given prompt–LLM pairing. This section examines (1) the best-performing prompts across models, (2) the best-performing prompts across tasks, and (3) the best-performing models across tasks. Unsurprisingly, we find that the optimal configuration varies with context.

\paragraph{Model-Specific Best Prompting Strategies}
Table \ref{tab:BestPromptByModelTask} (in Appendix A) and Figure \ref{fig:bestPropLLMTask} report the best-performing prompting method for each LLM. They reveal distinct model-specific preferences; no universally optimal strategy exists. With CoT-few-shot prompting, GPT-4o achieved the highest overall accuracy (0.907); thus, structured reasoning and exemplar calibration seem to pair well for large-scale models. By contrast, GPT-4o-mini performed best with CoT prompting and scored the highest F1 among compact models (0.830). As such, smaller architectures may benefit disproportionately from explicit reasoning instructions (to offset limited representational capacity). Claude-3.5-Haiku, DeepSeek-Chat-v3, and Gemini-2.5-Flash performed optimally with zero-shot prompting, reaching competitive accuracies (0.876–0.901) and consistently high recall (0.850–0.857). These models appear to generalize effectively without additional scaffolding. However, their lower precision (0.713–0.774) indicates a higher false-positive risk (thereby limiting their precision-sensitive applications). Llama-4-Maverick aligned best with CoT-few-shot (accuracy = 0.887; F1 = 0.774), where reasoning and exemplar guidance came together to counteract Llama's recall–precision imbalance.

In sum, our findings suggest a bifurcation: reasoning-oriented prompts (CoT, CoT-few-shot) enhance performance in GPT models and Llama-4, while zero-shot remains most effective for Claude, DeepSeek, and Gemini. Importantly, for prompting to prove effective, model architecture and the prompt's cognitive scaffolding must align; such harmony matters more than prompt complexity.

\begin{figure}[!htbp]
  \centering
  \includegraphics[width=0.7\textwidth]{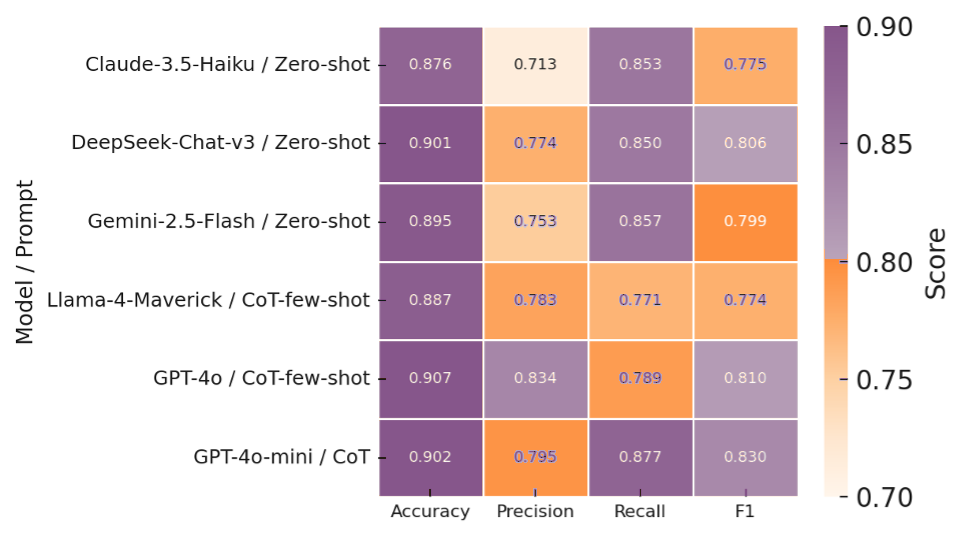}
  \caption{Best Performing Prompt Method for Each LLM (tasks classifications)}
  \label{fig:bestPropLLMTask}
\end{figure}

\begin{figure}[!htbp]
  \centering
  \includegraphics[width=0.7\textwidth]{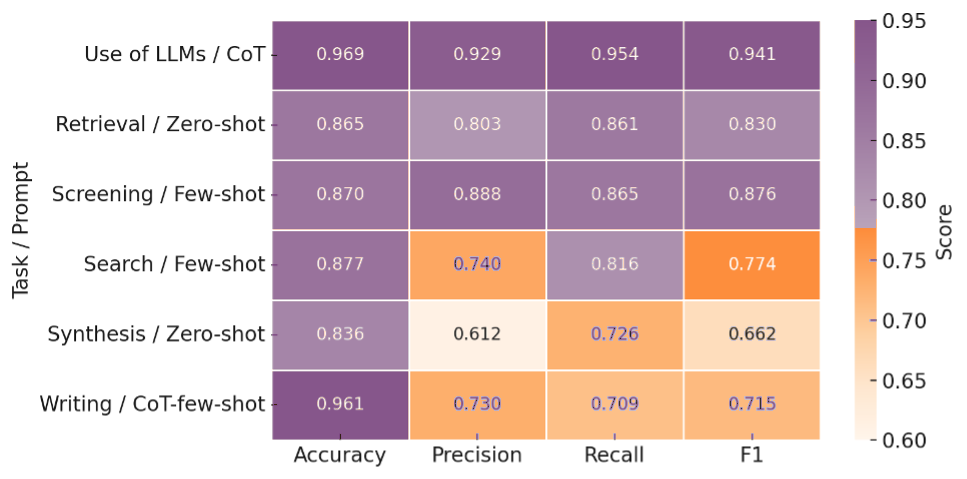}
  \caption{Best-Performing Prompt Method For Each Task}
  \label{fig:bestPromTask}
\end{figure}

\paragraph{Classification-Task-Specific Optimal Prompting Strategies}
As shown in Table \ref{tab:BestPromptTask} (Appendix A) and Figure \ref{fig:bestPromTask}, optimal prompting strategy varies by task. For LLM-use detection, CoT prompting achieved near-perfect results (accuracy = 0.969; F1 = 0.941); accordingly, structured reasoning matters very much to this task. In retrieval, zero-shot prompting performed best, yielding strong recall (0.861) and competitive accuracy (0.865); however, its moderate precision (0.803) reflects inclusivity at the expense of selectivity. Screening benefited most from few-shot prompting, which balanced high precision (0.888) with strong recall (0.865): annotated exemplars minimized false positives and negatives alike. Similarly, few-shot optimized searching but precision remained lower (0.740): broad retrieval patterns improved recall (0.816) while reducing specificity. Overall, synthesis remains the weakest task. Zero-shot prompting yielded low precision (0.612) and F1 (0.662) because abstract integration requires structured reasoning or richer contextual input. Writing, by contrast, performed best under CoT-few-shot conditions, achieving high accuracy (0.961) but only moderate precision (0.730); these metrics reflect frequent over-inclusion despite reliable detection.

Generally, reasoning-oriented prompts (CoT, CoT-few-shot) excel in conceptually demanding tasks (LLM detection, writing classification). Screening and searching classification benefit most from example-based prompts (few-shot) because such tasks require balanced trade-offs. Zero-shot prompting works well for retrieval but remains inadequate for synthesis classification because it requires deeper contextual reasoning.

\begin{figure}[!htbp]
  \centering
  \includegraphics[width=0.7\textwidth]{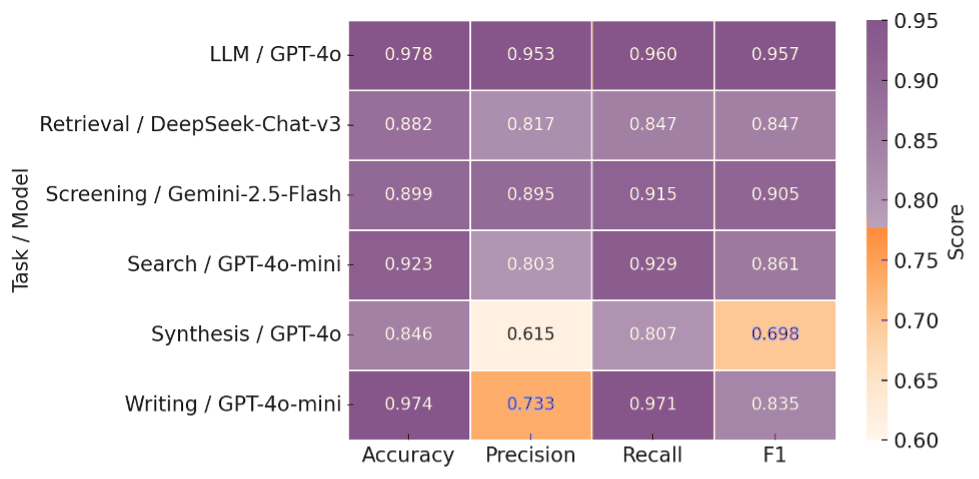}
  \caption{Best-Performing LLM For Each Task}
  \label{fig:bestModelTask}
\end{figure}

\paragraph{Classification-Task-Specific Best LLMs}
As seen in Table \ref{tab:bestLLMTask} (Appendix A) and Figure \ref{fig:bestModelTask}, the best-performing LLM systematically varies by SLR stage/LLM-use detection. GPT-4o achieved the highest performance in LLM-use detection (accuracy = 0.978; F1 = 0.957) because it could capture explicit lexical markers of model usage. For retrieval classification, DeepSeek-Chat-v3 provided the best balance (precision = 0.847; recall = 0.847; F1 = 0.847); it was broadly inclusive but did not excessively count false positives. Gemini-2.5-Flash (F1 = 0.905) best supported screening classification; it combined high precision (0.895) with strong recall (0.915), benefiting borderline inclusion decisions. For searching classification, GPT-4o-mini delivered the strongest results (accuracy = 0.923; recall = 0.929), though its lower precision (0.803) suggests a bias toward over-retrieval. GPT-4o-mini also performed best in writing classification (accuracy = 0.974; recall = 0.971) but was only moderately precise (0.733) and tended toward over-inclusivity. Synthesis classification was weakest because it involved semantically complex content; GPT-4o led this task (F1 = 0.698), driven by recall (0.807) but limited by low precision (0.615).

In aggregate, clear patterns emerge: GPT-4o and GPT-4o-mini dominate tasks anchored by explicit lexical cues (LLM use, searching, writing), while Gemini-2.5-Flash and DeepSeek-Chat-v3 excel in tasks requiring interpretive judgment (screening, retrieval). Across models, synthesis classification remains a persistent limitation: to improve precision in abstract-reasoning tasks, more advanced approaches are required.

\subsubsection{Confidence Intervals for Task Data}
As shown in Table \ref{tab:tasks-prompts-models-ci} (Appendix A), accuracy CIs were uniformly narrow, reflecting large denominators and metric stability. Precision and recall produced wider intervals, especially in tasks with fewer positives (e.g., synthesis, writing) because class imbalance magnifies variability. F1 CIs, derived via bootstrap resampling, were broader than accuracy CIs but accurately reflected uncertainty. In brief, stable tasks (like screening) tend to yield tighter intervals, while conceptually complex and data-sparse tasks tend to yield broader intervals; thus, we should cautiously interpret small performance differences.

\section{Discussion}
In this section, we (1) synthesize our main findings, (2) outline cost implications, (3) acknowledge key limitations, and (4) suggest avenues for future research. We use accuracy as a measure of overall correctness, and we use F1 to take a balanced look at precision and recall. In doing so, we clarify trade-offs inherent in different prompting strategies and model choices.

\subsection{Classification of Relevance to SLR Automation}
To answer RQ1, we now discuss how prompt design and model choice individually and jointly affect screening performance. To begin, by evaluating prompting strategies and model-specific effects in relevance classification, we yield several practical insights. 

\textbf{First}, CoT-few-shot prompting emerges as the most reliable baseline, combining high precision with a strong F1. It represents a pragmatic “default” for workflows that must minimize false positives without incurring substantial recall losses.

\textbf{Second}, zero-shot prompting works well in recall-critical contexts (like early-stage screening) where under-inclusion would be more detrimental than over-inclusion. A layered approach offers practical balance between sensitivity and efficiency: first applying zero-shot for broad inclusivity, followed by CoT-few-shot or few-shot for refinement.

\textbf{Third}, self-reflection prompting underperforms by over-including false positives without commensurate recall gains. Self-reflection also exhibits high cross-model variance, which suggests strong prompt–model interaction. Mechanistically, reflection encourages justificatory reasoning weakly anchored to explicit decision rules; without exemplar grounding, a model can rationalize initial errors without correcting them. By contrast, CoT-few-shot stabilizes decisions by pairing stepwise reasoning with concrete examples, thus yielding a better precision–recall balance.

\begin{figure}[!htbp]
  \centering
  \includegraphics[width=0.7\textwidth]{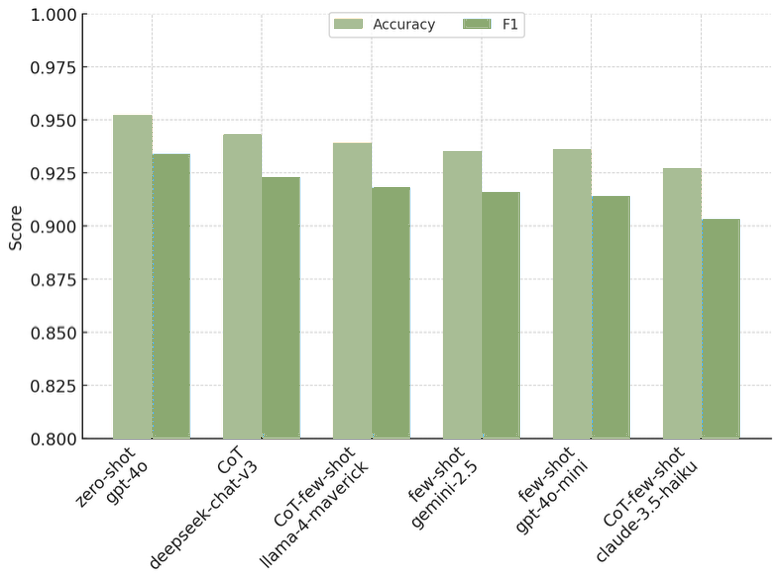}
  \caption{Best Performing Prompt Type by Model (relevance classification)}
  \label{fig:accuracyF1Rel}
\end{figure}

\textbf{Finally}, our results highlight the importance of model–prompt alignment. GPT-4o and DeepSeek consistently deliver robust all-round performance, while Gemini and Llama become competitive when paired with precision-enhancing prompts. Thus, for optimal performance, prompt design should be tailored to model strengths without relying on a universally superior configuration (Figure\ref{fig:accuracyF1Rel}).

\subsection{Classification Tasks Corresponding to SLR Stages and LLM Use}
\begin{figure}[!htbp]
  \centering
  \includegraphics[width=1.0\textwidth]{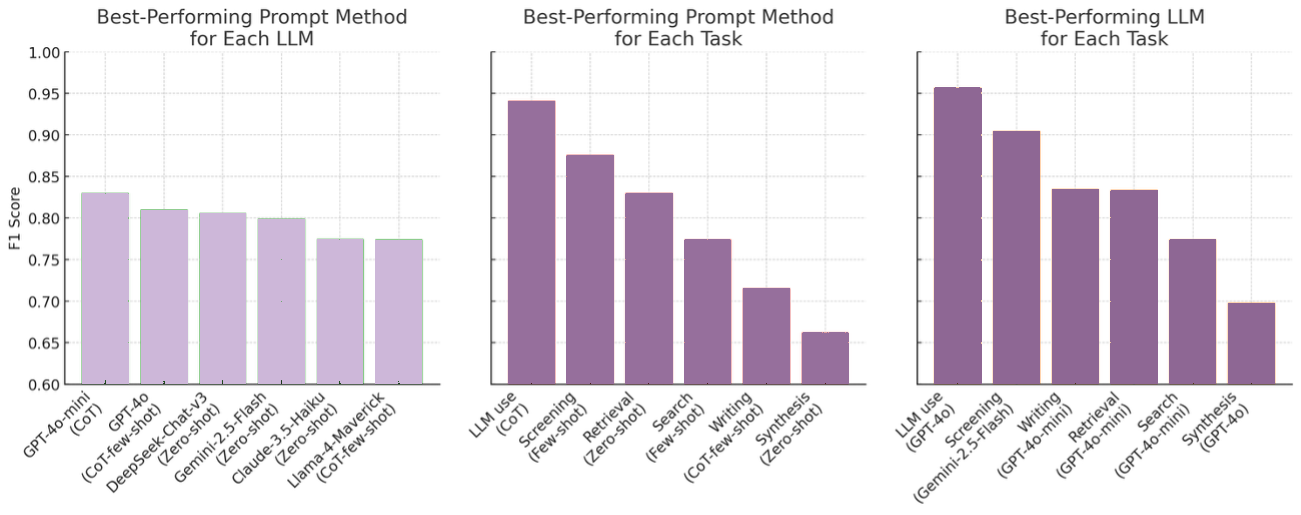}
  \caption{Comparative Performance of Best Prompts and LLMs Across Tasks}
  \label{fig:comProLLMTask}
\end{figure}

We now continue to answer RQ1 by moving from relevance classification to task classification. Across task dimensions, prompting strategies systematically influence the recall–precision balance: zero-shot maximizes inclusivity, while CoT-few-shot achieves the most reliable equilibrium. GPT-4o and GPT-4o-mini perform most consistently across tasks, while Gemini and DeepSeek remain competitive but more prompt-sensitive. Claude and Llama emphasize recall but require structured prompting to attain adequate precision (Figure \ref{fig:comProLLMTask}, left panel).

Performance varies considerably by task. Screening and retrieval classification yield the highest accuracy, reflecting clearer operational definitions and stronger dataset representation. By contrast, conceptual complexity and limited data constrain synthesis and writing classification; therefore, such tasks remain most challenging. LLM-use detection outperforms all other categories, achieving very high precision and recall. We may attribute this success to distinctive lexical markers (e.g., “ChatGPT” and “large language models”) (Figure \ref{fig:comProLLMTask}, mid \& right panels). Therefore, no universally optimal configuration exists: outcomes depend on interactions among task characteristics, model behaviour, and prompt design. A layered strategy (using zero-shot to maximize recall before using CoT-few-shot to maximize precision) emerges as a pragmatic semi-automation approach.

In sum, LLMs already provide reliable support for screening classification, retrieval classification, and LLM-use detection. However, more abstract tasks (synthesis classification, writing classification) require additional methodological innovation and larger datasets before they can reach comparable automation levels.

\subsection{Cost Analysis}
To answer RQ2, we begin by evaluating different model–prompt combinations' cost-performance implications. In Table \ref{tab:relevant-cost-f1}, we report costs of relevance classification per 1,000 abstracts across model–prompt pairings, alongside corresponding F1 scores. We find that across model-prompt pairings, absolute-dollar cost differences remain substantial. In particular, GPT-4o-Mini incurred very low costs across all prompt types (holding costs equal, GPT-4o-Mini's F1 was highest for 4/5 prompts). Llama-4-Maverick only led cost savings when paired with CoT-few-shot (+0.008), relative to larger models (e.g., GPT-4o, Claude-3.5-Haiku). Structured prompts (CoT, CoT-few-shot) delivered acceptable F1 on GPT-4o-Mini, despite increased token usage. Accordingly, prompts can be prototyped by pairing GPT-4o-Mini with CoT/CoT-few-shot (and, where useful, brief self-reflection) to inexpensively explore instruction variants (before using a higher-capacity model to conduct targeted validation). Furthermore, to take a first pass over a large volume of abstracts, low-cost models (e.g., GPT-4o-Mini, DeepSeek-Chat-V3) can be paired with structured prompts (CoT, CoT-few-shot) to favourably balance cost and performance; they should only escalate borderline and contentious cases for adjudication by a stronger model (e.g., GPT-4o). This staged strategy preserves most attainable accuracy and lowers absolute expenditure.

\begin{table}[!htbp]
\centering
\caption{Cost of Relevance Classification Per 1000 Abstracts (Tokens are totals; Cost in USD)}
\label{tab:relevant-cost-f1}
\resizebox{\textwidth}{!}{%
\begin{tabular}{l rr *{6}{rr}}
\toprule
& \multicolumn{2}{c}{\textbf{Tokens}} 
& \multicolumn{2}{c}{\makecell[c]{\textbf{Claude-3.5-}\\\textbf{Haiku}}}
& \multicolumn{2}{c}{\makecell[c]{\textbf{DeepSeek-}\\\textbf{Chat-V3}}}
& \multicolumn{2}{c}{\makecell[c]{\textbf{Gemini-2.5-}\\\textbf{Flash}}}
& \multicolumn{2}{c}{\makecell[c]{\textbf{GPT-}\\\textbf{4o}}}
& \multicolumn{2}{c}{\makecell[c]{\textbf{GPT-4o}\\\textbf{Mini}}}
& \multicolumn{2}{c}{\makecell[c]{\textbf{Llama-4-}\\\textbf{Maverick}}} \\
\cmidrule(lr){2-3}\cmidrule(lr){4-5}\cmidrule(lr){6-7}\cmidrule(lr){8-9}\cmidrule(lr){10-11}\cmidrule(lr){12-13}\cmidrule(lr){14-15}
\textbf{Prompt Type} & \textbf{Input} & \textbf{Output} 
& \textbf{Cost} & \textbf{F1}
& \textbf{Cost} & \textbf{F1}
& \textbf{Cost} & \textbf{F1}
& \textbf{Cost} & \textbf{F1}
& \textbf{Cost} & \textbf{F1}
& \textbf{Cost} & \textbf{F1} \\
\midrule
CoT             & 433{,}460 & 85{,}477  & \$0.69 & 0.884 & \$0.18 & 0.923 & \$0.34 & 0.887 & \$1.94 & 0.931 & \textbf{\$0.12} & 0.910 & \textbf{\$0.12} & 0.894 \\
CoT-few-shot    & 874{,}060 & 85{,}113  & \$1.04 & 0.903 & \$0.28 & 0.921 & \$0.48 & 0.908 & \$3.04 & 0.918 & \textbf{\$0.18} & 0.910 & \textbf{\$0.18} & 0.918 \\
Few-shot        & 859{,}460 & 38{,}060  & \$0.84 & 0.898 & \$0.24 & 0.918 & \$0.35 & 0.916 & \$2.53 & 0.927 & \textbf{\$0.15} & 0.914 & \textbf{\$0.15} & 0.897 \\
Self-reflection & 656{,}120 & 112{,}590 & \$0.98 & 0.811 & \$0.25 & 0.888 & \$0.48 & 0.863 & \$2.77 & 0.882 & \textbf{\$0.17} & 0.739 & \textbf{\$0.17} & 0.733 \\
Zero-shot       & 419{,}060 & 38{,}080  & \$0.49 & 0.887 & \$0.13 & 0.920 & \$0.22 & 0.891 & \$1.43 & 0.934 & \textbf{\$0.09} & 0.896 & \textbf{\$0.09} & 0.843 \\
\bottomrule
\end{tabular}
}%
\end{table}

\subsection{Limitations and Future Directions}
Domain scope is our study's primary limitation. We curated our dataset to SLR automation, so it contains salient topic-specific lexical markers (e.g., ``systematic review'', ``screening'', ``automation''). These may (1) invite shortcut learning and (2) overestimate generalizability to out-of-domain contexts that contain subtler signals (e.g., clinical meta-analyses guided by PICO formulations, social policy, and ecology). Consequently, absolute performance estimates may not transfer to domains with non-explicit inclusion criteria; our findings likely remain most valid for corpora that are topically structured like SLR automation.

We evaluated manually refined prompting strategies but not systematically iterative methods, meta-prompting methods \cite{liu2023selfreflection, zhang2025metapromptingaisystems}, or declarative prompt optimization \cite{susnjak2025compiling} (which might enable more effective, efficient prompt refinement). We did not test retrieval-augmented generation (RAG), although RAG may improve precision at a fixed, high recall and yield more stable, transferable predictions across heterogeneous abstracts by grounding decisions in task-relevant passages. In addition, class balance and prevalence varied across Level-2 tasks (e.g., we encountered relatively few ``writing'' and ``synthesis'' papers); this may have increased uncertainty and depressed F1 in abstraction-heavy tasks. Also, despite consensus labelling, manual annotation may have introduced residual subjectivity, especially in under-represented categories (like synthesis and writing). Finally, our evaluation emphasizes quantitative metrics without directly assessing qualitative dimensions (e.g., interpretability and trustworthiness).

To overcome domain-scope constraints and to test transportability, future work should build cross-domain benchmarks that span settings with subtler signals (e.g., PICO-guided clinical meta-analyses, social policy, and ecology); performance can be reported under distribution shift using fixed-recall operating points and workload-reduction curves. Later studies may also evaluate (1) retrieval-augmented pipelines (that ground decisions in methods-level evidence) with (2) systematic prompt development (iterative/meta-prompting and declarative optimization) as domain-adaptation mechanisms (not merely accuracy boosters). Class imbalance warrants stratified sampling, prevalence reporting, and cost-sensitive/abstention-aware thresholds, complemented by calibration analyses and human-in-the-loop audits. Finally, reproducibility and external validity (across heterogeneous screening contexts) can be strengthened via (1) qualitative error analysis, (2) user studies of interpretability, (3) preregistered protocols, and (4) openly released code, prompts, indexes, and annotation rubrics.

\subsection{Practical Recommendations}
Based on our task-specific evaluation of prompting strategy and LLM performance, several practical recommendations emerge for LLM integration into SLR workflows:

\begin{itemize}[noitemsep, topsep=0pt]
\setlength\itemsep{4pt}      
\item \textbf{Account for task complexity.} Supply detailed inclusion/exclusion criteria and contextual framing to scaffold reasoning when targets are less lexically anchored.
\setlength\itemsep{4pt}      
\item \textbf{Adopt CoT-few-shot as a baseline.} CoT-few-shot consistently offers the best precision–recall balance for general screening; thus, it should be your default starting point.
\setlength\itemsep{4pt}      
\item \textbf{Use zero-shot in recall-critical stages.} In early triage (where missed inclusions cost more), apply zero-shot to maximize recall before refining with CoT-few-shot or few-shot to reduce false positives.
\setlength\itemsep{4pt}      
\item \textbf{Align prompts with model characteristics.} GPT-4o and DeepSeek remain robust across prompting strategies, while Gemini and Llama benefit most from precision-enhancing prompts (e.g. CoT-few-shot).
\setlength\itemsep{4pt}      
\item \textbf{Prefer reasoning-based prompts for abstract tasks.} Use CoT or CoT-few-shot for classifications that demand deeper contextual reasoning. 
\setlength\itemsep{4pt}      
\item \textbf{Cost-aware prompt development}. Prototype on a small calibration set using a low-cost model (e.g., GPT-4o-mini), applying CoT or CoT-few-shot to tune instructions. Validate on a higher-capacity model (e.g., GPT-4o) only if doing so would deliver a measurable performance gain for your intended use. 
\setlength\itemsep{4pt}      
\item \textbf{First-pass at scale.} When confronted with a large volume of abstracts, combine a low-cost model (e.g., GPT-4o-mini or DeepSeek-Chat-V3 where available) with structured prompts (CoT/CoT-few-shot) to achieve a favourable cost–performance balance. Escalate only borderline/contentious cases to a stronger model. 
\end{itemize}

Generally, clearly defined instructions and contextual cues lead to effective prompting, while LLM-specific and task-specific characteristics guide strategy selection.

\section{Conclusion}
According to our cross-model evaluation, prompting strategy, model choice, and task type jointly determine screening effectiveness. With regard to prompting strategy: (1) CoT-few-shot offers the most reliable precision–recall balance, (2) zero-shot maximizes recall for high-sensitivity passes, and (3) self-reflection performs worst because it exhibits high model-specific variance and over-inclusivity. With regard to model choice: GPT-4o and DeepSeek deliver strong all-round results, while GPT-4o-Mini performs competitively at a markedly lower cost. Our relevance-classification cost analysis confirms substantial absolute differences among model–prompt pairings. Thus, it supports a staged, cost-aware workflow: researchers should deploy low-cost models with structured prompts for first-pass screening before routing borderline cases to higher-capacity models. These conclusions apply primarily to SLR-automation corpora with salient lexical cues; they may or may not transport well to PICO-anchored/less explicit domains. Future work should include cross-domain external validation and evaluation of retrieval-augmented and prompt-optimization methods under human-in-the-loop conditions with prospective workload and cost tracking.



\begingroup
\raggedbottom

\clearpage   
\appendix

\section{Additional Data Tables and Brief Analysis}\label{app:A}

\subsection{Evaluation Metrics for Prompt–Model Combinations in Relevance Classification}
Table \ref{tab:PromtModelCombinaRelevan} reports relevance-classification performance (Accuracy, Precision, Recall, F1 with 95\% CIs) for six LLMs across five prompting strategies (CoT, CoT-few-shot, Few-shot, Self-reflection, Zero-shot). Patterns indicate strong model–prompt interactions: GPT-4o attains the highest F1 under Zero-shot (0.934) and Few-shot (0.927), while DeepSeek leads under CoT-few-shot (0.921). CoT-few-shot generally shifts the operating point toward higher precision at some cost to recall (e.g., GPT-4o Precision 0.952 with Recall 0.886), whereas Zero-shot and CoT favour very high recall (often $\geq$\,0.98 for Gemini and Llama) with correspondingly lower precision and middling F1. Self-reflection underperforms relative to simpler prompts (notably for GPT-4o-Mini and Llama), suggesting limited benefit for this task. In short, if minimising false positives is paramount, CoT-few-shot is preferable; if maximising inclusivity (recall) is the goal, Zero-shot/CoT are more suitable, with GPT-4o consistently the most robust.

\renewcommand{\thetable}{A.\arabic{table}}

\setcounter{table}{0}  

\begin{table}[!htbp]
\centering
\caption{Evaluation Metrics for Prompt–Model Combinations in Relevance Classification}
\label{tab:PromtModelCombinaRelevan}
\normalsize
\begin{tabular}{
  >{\centering\arraybackslash}p{5mm}    
  >{\raggedright\arraybackslash}p{29mm} 
  >{\centering\arraybackslash}m{30mm}   
  >{\centering\arraybackslash}m{30mm}   
  >{\centering\arraybackslash}m{30mm}   
  >{\centering\arraybackslash}m{30mm}   
}
\toprule
PT & Model & Accuracy (95\% CI) & Precision (95\% CI) & Recall (95\% CI) & F1 (95\% CI) \\
\midrule
\multirow{6}{*}{\rotv{CoT}}
  & Claude-3.5-Haiku   & 0.907 [0.891, 0.923] & 0.808 [0.775, 0.840] & 0.976 [0.961, 0.989] & 0.884 [0.863, 0.904] \\
  & Deepseek-Chat-V3   & 0.943 [0.931, 0.955] & 0.901 [0.875, 0.925] & 0.945 [0.924, 0.964] & 0.923 [0.905, 0.938] \\
  & Gemini-2.5-Flash   & 0.910 [0.895, 0.925] & 0.804 [0.772, 0.836] & 0.990 [0.980, 0.998] & 0.887 [0.867, 0.907] \\
  & GPT-4o             & \textbf{0.950} [0.938, 0.961] & \textbf{0.919} [0.894, 0.941] & 0.943 [0.923, 0.962] & \textbf{0.931} [0.914, 0.946] \\
  & GPT-4o-Mini        & 0.934 [0.921, 0.947] & 0.885 [0.857, 0.911] & 0.937 [0.914, 0.958] & 0.910 [0.891, 0.928] \\
  & Llama-4-Maverick   & 0.916 [0.901, 0.930] & 0.814 [0.783, 0.844] & \textbf{0.992} [0.983, 0.998] & 0.894 [0.875, 0.913] \\
\cmidrule(lr){1-6}
\multirow{6}{*}{\rotv{CoT-few-shot}}
  & Claude-3.5-Haiku   & 0.927 [0.913, 0.941] & 0.859 [0.829, 0.887] & 0.953 [0.933, 0.970] & 0.903 [0.884, 0.921] \\
  & Deepseek-Chat-V3   & \textbf{0.944} [0.932, 0.956] & 0.931 [0.907, 0.953] & 0.910 [0.885, 0.934] & \textbf{0.921} [0.902, 0.938] \\
  & Gemini-2.5-Flash   & 0.928 [0.914, 0.941] & 0.840 [0.810, 0.869] & \textbf{0.988} [0.977, 0.996] & 0.908 [0.889, 0.925] \\
  & GPT-4o             & 0.943 [0.931, 0.955] & \textbf{0.952} [0.932, 0.971] & 0.886 [0.858, 0.915] & 0.918 [0.899, 0.936] \\
  & GPT-4o-Mini        & 0.938 [0.925, 0.951] & 0.943 [0.921, 0.963] & 0.880 [0.850, 0.908] & 0.910 [0.891, 0.929] \\
  & Llama-4-Maverick   & 0.939 [0.926, 0.951] & 0.879 [0.850, 0.906] & 0.961 [0.943, 0.977] & 0.918 [0.900, 0.935] \\
\cmidrule(lr){1-6}
\multirow{6}{*}{\rotv{Few-shot}}
  & Claude-3.5-Haiku   & 0.922 [0.907, 0.935] & 0.838 [0.807, 0.867] & 0.967 [0.951, 0.982] & 0.898 [0.878, 0.917] \\
  & Deepseek-Chat-V3   & 0.942 [0.930, 0.954] & \textbf{0.924} [0.899, 0.946] & 0.913 [0.888, 0.937] & 0.918 [0.900, 0.935] \\
  & Gemini-2.5-Flash   & 0.935 [0.922, 0.949] & 0.860 [0.830, 0.888] & \textbf{0.980} [0.967, 0.992] & 0.916 [0.898, 0.933] \\
  & GPT-4o             & \textbf{0.948} [0.935, 0.959] & 0.922 [0.897, 0.945] & 0.933 [0.910, 0.954] & \textbf{0.927} [0.910, 0.943] \\
  & GPT-4o-Mini        & 0.936 [0.923, 0.948] & 0.880 [0.850, 0.906] & 0.951 [0.931, 0.970] & 0.914 [0.895, 0.931] \\
  & Llama-4-Maverick   & 0.919 [0.904, 0.933] & 0.827 [0.796, 0.857] & 0.980 [0.966, 0.991] & 0.897 [0.877, 0.915] \\
\cmidrule(lr){1-6}
\multirow{6}{*}{\rotv{Self-reflection}}
  & Claude-3.5-Haiku   & 0.839 [0.815, 0.861] & 0.708 [0.668, 0.747] & 0.949 [0.925, 0.970] & 0.811 [0.782, 0.838] \\
  & Deepseek-Chat-V3   & \textbf{0.918} [0.903, 0.932] & \textbf{0.863} [0.834, 0.892] & 0.914 [0.888, 0.938] & \textbf{0.888} [0.867, 0.908] \\
  & Gemini-2.5-Flash   & 0.889 [0.872, 0.906] & 0.774 [0.741, 0.807] & \textbf{0.974} [0.959, 0.987] & 0.863 [0.841, 0.883] \\
  & GPT-4o             & 0.910 [0.895, 0.926] & 0.837 [0.806, 0.868] & 0.931 [0.907, 0.952] & 0.882 [0.860, 0.902] \\
  & GPT-4o-Mini        & 0.763 [0.741, 0.786] & 0.609 [0.575, 0.644] & 0.939 [0.916, 0.959] & 0.739 [0.712, 0.766] \\
  & Llama-4-Maverick   & 0.758 [0.735, 0.780] & 0.606 [0.570, 0.640] & 0.929 [0.905, 0.951] & 0.733 [0.705, 0.760] \\
\cmidrule(lr){1-6}
\multirow{6}{*}{\rotv{Zero-shot}}
  & Claude-3.5-Haiku   & 0.910 [0.895, 0.924] & 0.804 [0.772, 0.836] & 0.988 [0.977, 0.996] & 0.887 [0.866, 0.906] \\
  & Deepseek-Chat-V3   & 0.941 [0.929, 0.954] & 0.898 [0.872, 0.923] & 0.943 [0.922, 0.962] & 0.920 [0.903, 0.937] \\
  & Gemini-2.5-Flash   & 0.913 [0.898, 0.927] & 0.810 [0.778, 0.840] & 0.990 [0.980, 0.998] & 0.891 [0.871, 0.909] \\
  & GPT-4o             & \textbf{0.952} [0.940, 0.963] & \textbf{0.921} [0.897, 0.944] & 0.947 [0.926, 0.967] & \textbf{0.934} [0.918, 0.949] \\
  & GPT-4o-Mini        & 0.920 [0.906, 0.934] & 0.839 [0.809, 0.869] & 0.962 [0.944, 0.978] & 0.896 [0.876, 0.915] \\
  & Llama-4-Maverick   & 0.867 [0.848, 0.884] & 0.729 [0.696, 0.762] & \textbf{0.998} [0.994, 1.000] & 0.843 [0.820, 0.864] \\
\bottomrule
\end{tabular}
\vspace{0.3em}
\begin{minipage}{\textwidth}
\small\emph{Note.} All metrics are reported with 95\% confidence intervals in brackets. 
Boldface highlights the best-performing \emph{point estimate} within each prompt type and metric (CIs not bolded). 
\textbf{PT} = Prompt Type (CoT, CoT-few-shot, Few-shot, Self-reflection, Zero-shot).
\end{minipage}
\end{table}

\vspace{2em}
\subsection{Best Performing Prompt Types By Model for Relevance Classification}
Table \ref{tab:BestPromptByModel} shows evaluation metrics of best performing prompt types by model for relevance classification. GPT-4o attains the top result under Zero-shot (Acc 0.952, Prec 0.921, F1 0.934), showing strong out-of-the-box calibration. Peaks vary by model, Deepseek-Chat-V3 with CoT (F1 0.923), Llama-4-Maverick with CoT-few-shot (F1 0.918), indicating prompt–model interactions. Gemini-2.5’s best Few-shot setting prioritizes recall (0.980) over precision (0.860), whereas GPT-4o’s Zero-shot is more precision-balanced. In brief, best-of-prompt F1s cluster tightly (0.903–0.934), with a modest but consistent frontier-model edge.

\vspace{2em}

\begin{table}[!htbp]
\centering
\caption{Best Performing Prompt Types by Model (relevance classification)}
\label{tab:BestPromptByModel}
\begin{tabular}{llcccc}
\toprule
\textbf{Model} & \textbf{Prompt type} & \textbf{Accuracy} & \textbf{Precision} & \textbf{Recall} & \textbf{F1} \\
\midrule
Gpt-4o             & Zero-shot      & \textbf{0.952} & \textbf{0.921} & 0.947 & \textbf{0.934} \\
Deepseek-Chat-V3   & CoT            & 0.943 & 0.901 & 0.945 & 0.923 \\
Llama-4-Maverick   & CoT-few-shot   & 0.939 & 0.879 & 0.961 & 0.918 \\
Gemini-2.5         & Few-shot       & 0.935 & 0.860 & \textbf{0.980} & 0.916 \\
Gpt-4o-Mini        & Few-shot       & 0.936 & 0.880 & 0.951 & 0.914 \\
Claude-3.5-Haiku   & CoT-few-shot   & 0.927 & 0.859 & 0.953 & 0.903 \\
\bottomrule
\end{tabular}
\end{table}

\vspace{4em}

\subsection{Best-Performing Prompt Method for Each LLM of Tasks Classifications}
Table \ref{tab:BestPromptByModelTask} lists, for each LLM, the prompt method that yielded its best task-classification performance and the associated metrics. Zero-shot is optimal for Claude-3.5-Haiku, DeepSeek-Chat-v3, and Gemini-2.5-Flash, reflecting strong transfer with minimal prompt design but a recall-leaning profile (e.g., Claude Recall 0.853 vs. Precision 0.713). Structured prompting benefits others: Llama-4-Maverick and GPT-4o peak under CoT-few-shot, with GPT-4o achieving the highest Accuracy (0.907) and Precision (0.834), while GPT-4o-mini attains the highest Recall (0.877) and F1 (0.830) under plain CoT. The pattern suggests that frontier or architecture-sensitive models exploit reasoning structure for better calibration, whereas several models still reach their best with simple zero-shot prompting.

\vspace{2em}

\begin{table}[!htbp]
\centering
\caption{Best-Performing Prompt Method for Each LLM (tasks classifications)}
\label{tab:BestPromptByModelTask}
\begin{tabular}{lccccc}
\toprule
\textbf{Model} & \textbf{Best Prompt} & \textbf{Accuracy} & \textbf{Precision} & \textbf{Recall} & \textbf{F1} \\
\midrule
Claude-3.5-Haiku   & Zero-shot     & 0.876 & 0.713 & 0.853 & 0.775 \\
DeepSeek-Chat-v3   & Zero-shot     & 0.901 & 0.774 & 0.850 & 0.806 \\
Gemini-2.5-Flash   & Zero-shot     & 0.895 & 0.753 & 0.857 & 0.799 \\
Llama-4-Maverick   & CoT-few-shot  & 0.887 & 0.783 & 0.771 & 0.774 \\
GPT-4o             & CoT-few-shot  & \textbf{0.907} & \textbf{0.834} & 0.789 & 0.810 \\
GPT-4o-mini        & CoT           & 0.902 & 0.795 & \textbf{0.877} & \textbf{0.830} \\
\bottomrule
\end{tabular}
\end{table}

\pagebreak

\subsection{Best-Performing Prompt Method For Each Task of Automation Tasks Classifications}
Table \ref{tab:BestPromptTask} identifies, for each task, the prompt that yields the best performance. CoT dominates \emph{LLM use} with the top scores across all metrics (Accuracy 0.969, Precision 0.929, Recall 0.954, F1 0.941), indicating prominent clues. \emph{Screening} and \emph{Searching} favour Few-shot, with Screening achieving the strongest F1 (0.876; Precision 0.888, Recall 0.865), suggesting exemplars calibrate decision boundaries effectively. \emph{Retrieval} and \emph{Synthesis} peak under Zero-shot (F1 0.830 and 0.662, respectively), implying broad generalisation without examples can help, though Synthesis remains the hardest (low Precision 0.612). \emph{Writing} prefers CoT-few-shot with very high Accuracy (0.961) but modest F1 (0.715), consistent with class imbalance or threshold sensitivity in generative judgements.

\begin{table}[!htbp]
\centering
\caption{Best-Performing Prompt Method For Each Task}
\label{tab:BestPromptTask}
\begin{tabular}{lccccc}
\toprule
\textbf{Task} & \textbf{Best Prompt} & \textbf{Accuracy} & \textbf{Precision} & \textbf{Recall} & \textbf{F1} \\
\midrule
LLM use        & CoT          & \textbf{0.969} & \textbf{0.929} & \textbf{0.954} & \textbf{0.941} \\
Retrieval      & Zero-shot    & 0.865 & 0.803 & 0.861 & 0.830 \\
Screening      & Few-shot     & 0.870 &\textbf{0.888} & \textbf{0.865} & \textbf{0.876} \\
Searching      & Few-shot     & 0.877 & 0.740 & 0.816 & 0.774 \\
Synthesis      & Zero-shot    & 0.836 & 0.612 & 0.726 & 0.662 \\
Writing        & CoT-few-shot & 0.961 & 0.730 & 0.709 & 0.715 \\
\bottomrule
\end{tabular}
\end{table}

\subsection{Best-Performing LLM for Each Task of Automation Tasks Classifications}
Table \ref{tab:bestLLMTask} names the top LLM for each task and its scores. Performance leadership is task-specific: GPT-4o leads \emph{LLM use} (Accuracy 0.978; Precision 0.953; F1 0.957) and \emph{Synthesis} (F1 0.698), though the latter’s low Precision (0.615) underscores task difficulty. \emph{Retrieval} is best with DeepSeek-Chat-v3 (balanced Precision/Recall 0.847), while \emph{Screening} favors Gemini-2.5-Flash (F1 0.905 from high Recall 0.915 with strong Precision 0.895). GPT-4o-mini excels in \emph{Searching} and \emph{Writing}, driven by very high Recall (0.929 and 0.971) but more modest Precision (0.803 and 0.733), making it well-suited for inclusive first-pass triage at the cost of more false positives. On balance, the winners differ by task, reflecting clear precision–recall trade-offs rather than a single universally best model.
\vspace{1em}
\begin{table}[!htbp]
\centering
\caption{Best-Performing LLM For Each Task}
\label{tab:bestLLMTask}
\begin{tabular}{lccccc}
\toprule
\textbf{Task} & \textbf{Best LLM} & \textbf{Accuracy} & \textbf{Precision} & \textbf{Recall} & \textbf{F1} \\
\midrule
LLM use        & GPT-4o             & \textbf{0.978} & \textbf{0.953} & 0.960 & \textbf{0.957}\\
Retrieval      & DeepSeek-Chat-v3   & 0.882 & 0.847 & 0.847 & 0.847 \\
Screening      & Gemini-2.5-Flash   & 0.899 & 0.895 & 0.915 & 0.905 \\
Searching      & GPT-4o-mini        & 0.923 & 0.803 & 0.929 & 0.861 \\
Synthesis      & GPT-4o             & 0.846 & 0.615 & 0.807 & 0.698 \\
Writing        & GPT-4o-mini        & 0.974 & 0.733 & \textbf{0.971} & 0.835 \\
\bottomrule
\end{tabular}
\end{table}

\subsection{Evaluation Metrics across Tasks, Prompt Types, and Models}
Table \ref{tab:tasks-prompts-models-ci} summarises Accuracy, Precision, Recall, and F1 (with 95\% CIs) across all tasks, prompt types, and models. Patterns indicate strong prompt–task interactions: Zero-shot and CoT tend to maximise recall (useful for inclusive first-pass screening), while CoT-few-shot shifts toward higher precision at some cost to recall. Easier tasks (\emph{LLM use}, \emph{Screening}) show uniformly high scores, whereas \emph{Synthesis} remains challenging with lower precision and F1. Among models, GPT-4o generally attains the strongest macro F1 with DeepSeek close behind; Gemini often posts the highest recall, and smaller/open models trail mainly on precision. In short, there is no single best prompt - choose recall-leaning prompts for sensitivity and CoT-few-shot when false positives are costlier.

\pagebreak

\setlength{\tabcolsep}{3pt}        
\renewcommand{\arraystretch}{1.05}  
\small                              
\FloatBarrier

\begin{longtable}{lllcccc}
\caption{Evaluation Mtrics across Tasks, Prompt Types, and Models}
\label{tab:tasks-prompts-models-ci}\\
\toprule
TK & PT & Model & Accuracy (95\% CI) & Precision (95\% CI) & Recall (95\% CI) & F1 (95\% CI) \\
\midrule
\endfirsthead

\toprule
TK & PT & Model & Accuracy (95\% CI) & Precision (95\% CI) & Recall (95\% CI) & F1 (95\% CI) \\
\midrule
\endhead

\midrule
\midrule

\multicolumn{7}{r}{\textit{Continued on next page}}\\
\endfoot

\bottomrule
\endlastfoot

\setTK{LLM} 

\multirow{24}{*}{\rotv{LLM}} 
  \ptblockstart
  \setPT{CoT}
& \multirow{6}{*}{\rotv{CoT}} & Claude-3.5-Haiku   & \textbf{0.975} [0.960, 0.987] & \textbf{0.951} [0.911, 0.984] & 0.951 [0.909, 0.985] & \textbf{0.951} [0.921, 0.976] \\
& & Deepseek-Chat-V3   & 0.972 [0.957, 0.986] & 0.944 [0.902, 0.979] & 0.944 [0.899, 0.980] & 0.944 [0.912, 0.971] \\
& & Gemini-2.5-Flash   & 0.972 [0.955, 0.986] & 0.918 [0.868, 0.962] & 0.976 [0.946, 1.000] & 0.946 [0.916, 0.972] \\
& & GPT-4o             & 0.968 [0.951, 0.982] & 0.930 [0.883, 0.971] & 0.945 [0.902, 0.983] & 0.938 [0.904, 0.966] \\
& & GPT-4o-Mini        & 0.963 [0.947, 0.980] & 0.936 [0.889, 0.975] & 0.921 [0.871, 0.964] & 0.929 [0.892, 0.959] \\
& & Llama-4-Maverick   & 0.966 [0.949, 0.980] & 0.892 [0.837, 0.940] & \textbf{0.984} [0.959, 1.000] & 0.936 [0.903, 0.964] \\

  \ptblocksep
  \ptblockstart
  
  \setPT{CoT-few-shot}
& \multirow{6}{*}{\rotv{CoT-few-shot}} & Claude-3.5-Haiku & 0.953 [0.934, 0.972] & 0.932 [0.880, 0.973] & 0.886 [0.826, 0.941] & 0.908 [0.867, 0.944] \\
& & Deepseek-Chat-V3 & 0.961 [0.943, 0.978] & \textbf{0.957} [0.917, 0.991] & 0.889 [0.829, 0.942] & 0.922 [0.883, 0.955] \\
& & Gemini-2.5-Flash & 0.972 [0.955, 0.986] & 0.931 [0.884, 0.970] & \textbf{0.960} [0.921, 0.992] & 0.945 [0.915, 0.971] \\
& & GPT-4o           & \textbf{0.978} [0.963, 0.990] & 0.953 [0.912, 0.985] & \textbf{0.960} [0.922, 0.992] & \textbf{0.957} [0.929, 0.980] \\
& & GPT-4o-Mini      & 0.963 [0.945, 0.980] & 0.943 [0.898, 0.982] & 0.913 [0.859, 0.960] & 0.927 [0.891, 0.959] \\
& & Llama-4-Maverick & 0.967 [0.951, 0.982] & 0.922 [0.874, 0.965] & 0.952 [0.911, 0.985] & 0.937 [0.904, 0.966] \\
  \ptblocksep
  \ptblockstart
   \setPT{few-shot}
& \multirow{6}{*}{\rotv{few-shot}} & Claude-3.5-Haiku & 0.949 [0.929, 0.967] & 0.939 [0.891, 0.980] & 0.856 [0.793, 0.917] & 0.895 [0.853, 0.934] \\
& & Deepseek-Chat-V3 & 0.968 [0.951, 0.982] & \textbf{0.951} [0.908, 0.984] & 0.921 [0.870, 0.964] & 0.935 [0.901, 0.964] \\
& & Gemini-2.5-Flash & 0.970 [0.953, 0.984] & \textbf{0.951} [0.909, 0.984] & 0.929 [0.877, 0.970] & 0.940 [0.907, 0.969] \\
& & GPT-4o           & \textbf{0.976} [0.961, 0.988] & 0.938 [0.894, 0.977] & 0.968 [0.934, 0.993] & \textbf{0.953} [0.924, 0.977] \\
& & GPT-4o-Mini      & 0.935 [0.911, 0.955] & 0.822 [0.757, 0.882] & 0.952 [0.911, 0.985] & 0.882 [0.837, 0.921] \\
& & Llama-4-Maverick & 0.919 [0.894, 0.943] & 0.761 [0.695, 0.826] & \textbf{0.992} [0.973, 1.000] & 0.861 [0.817, 0.902] \\
  \ptblocksep
  \ptblockstart
  \setPT{zero-shot}
& \multirow{6}{*}{\rotv{zero-shot}} & Claude-3.5-Haiku & 0.963 [0.945, 0.980] & 0.915 [0.862, 0.961] & 0.944 [0.900, 0.982] & 0.929 [0.893, 0.959] \\
& & Deepseek-Chat-V3 & \textbf{0.976} [0.961, 0.988] & \textbf{0.938} [0.892, 0.977] & 0.968 [0.934, 0.993] & \textbf{0.953} [0.924, 0.978] \\
& & Gemini-2.5-Flash & \textbf{0.976} [0.961, 0.988] & 0.932 [0.885, 0.971] & \textbf{0.976} [0.946, 1.000] & \textbf{0.953} [0.924, 0.978] \\
& & GPT-4o           & 0.961 [0.943, 0.978] & 0.897 [0.843, 0.945] & 0.961 [0.922, 0.992] & 0.928 [0.893, 0.958] \\
& & GPT-4o-Mini      & 0.921 [0.897, 0.943] & 0.778 [0.713, 0.840] & 0.969 [0.934, 0.993] & 0.863 [0.818, 0.902] \\
& & Llama-4-Maverick & 0.917 [0.892, 0.941] & 0.756 [0.690, 0.820] & \textbf{0.992} [0.973, 1.000] & 0.858 [0.814, 0.899] \\
\midrule
\setTK{Retrieval} 
\multirow{24}{*}{\rotv{Retrieval}} 
  \ptblockstart
& \multirow{6}{*}{\rotv{CoT}} & Claude-3.5-Haiku   & 0.813 [0.777, 0.847] & 0.695 [0.632, 0.755] & \textbf{0.892 [0.844, 0.935]} & 0.781 [0.734, 0.824] \\
& & Deepseek-Chat-V3   & 0.854 [0.822, 0.884] & 0.813 [0.756, 0.868] & 0.804 [0.746, 0.860] & 0.809 [0.764, 0.850] \\
& & Gemini-2.5-Flash   & 0.858 [0.828, 0.888] & 0.799 [0.743, 0.853] & 0.841 [0.788, 0.891] & \textbf{0.820} [0.776, 0.859] \\
& & GPT-4o             & \textbf{0.862} [0.832, 0.890] & \textbf{0.824} [0.768, 0.876] & 0.815 [0.757, 0.868] & 0.819 [0.775, 0.859] \\
& & GPT-4o-Mini        & 0.860 [0.828, 0.890] & 0.813 [0.758, 0.866] & 0.825 [0.768, 0.878] & 0.819 [0.776, 0.860] \\
& & Llama-4-Maverick   & 0.852 [0.819, 0.882] & 0.782 [0.725, 0.837] & 0.852 [0.801, 0.901] & 0.815 [0.773, 0.855] \\
  \ptblocksep
  \ptblockstart
& \multirow{6}{*}{\rotv{CoT-few-shot}} & Claude-3.5-Haiku & 0.820 [0.786, 0.854] & 0.740 [0.675, 0.802] & 0.802 [0.741, 0.858] & 0.770 [0.720, 0.816] \\
& & Deepseek-Chat-V3 & 0.840 [0.807, 0.872] & 0.799 [0.740, 0.855] & 0.778 [0.716, 0.835] & 0.788 [0.741, 0.832] \\
& & Gemini-2.5-Flash & 0.856 [0.824, 0.886] & 0.798 [0.741, 0.852] & \textbf{0.836} [0.782, 0.887] & 0.817 [0.774, 0.857] \\
& & GPT-4o           & \textbf{0.874} [0.844, 0.905] & \textbf{0.859} [0.807, 0.909] & 0.804 [0.749, 0.861] & \textbf{0.831} [0.788, 0.872] \\
& & GPT-4o-Mini      & 0.819 [0.785, 0.854] & 0.760 [0.701, 0.820] & 0.772 [0.712, 0.831] & 0.766 [0.717, 0.811] \\
& & Llama-4-Maverick & 0.827 [0.793, 0.859] & 0.756 [0.698, 0.815] & 0.809 [0.751, 0.862] & 0.781 [0.735, 0.825] \\
  \ptblocksep
  \ptblockstart
& \multirow{6}{*}{\rotv{few-shot}} & Claude-3.5-Haiku & 0.845 [0.812, 0.876] & 0.797 [0.739, 0.851] & 0.797 [0.738, 0.853] & 0.797 [0.751, 0.840] \\
& & Deepseek-Chat-V3 & 0.854 [0.822, 0.884] & 0.797 [0.742, 0.851] & 0.831 [0.777, 0.883] & 0.813 [0.770, 0.854] \\
& & Gemini-2.5-Flash & 0.852 [0.822, 0.883] & \textbf{0.809} [0.750, 0.863] & 0.809 [0.750, 0.862] & 0.809 [0.762, 0.850] \\
& & GPT-4o           & \textbf{0.858} [0.826, 0.888] & 0.787 [0.731, 0.843] & \textbf{0.862} [0.813, 0.910] & \textbf{0.823} [0.781, 0.863] \\
& & GPT-4o-Mini      & 0.840 [0.807, 0.872] & 0.755 [0.698, 0.812] & \textbf{0.862} [0.811, 0.910] & 0.805 [0.760, 0.845] \\
& & Llama-4-Maverick & 0.843 [0.809, 0.874] & 0.800 [0.740, 0.855] & 0.787 [0.727, 0.844] & 0.794 [0.745, 0.837] \\
  \ptblocksep
  \ptblockstart
& \multirow{6}{*}{\rotv{zero-shot}} & Claude-3.5-Haiku & 0.839 [0.804, 0.869] & 0.733 [0.675, 0.788] & \textbf{0.909} [0.867, 0.949] & 0.811 [0.769, 0.850] \\
& & Deepseek-Chat-V3 & \textbf{0.882} [0.854, 0.911] & \textbf{0.847} [0.796, 0.896] & 0.847 [0.794, 0.898] & \textbf{0.847} [0.806, 0.884] \\
& & Gemini-2.5-Flash & 0.860 [0.830, 0.890] & 0.806 [0.751, 0.860] & 0.836 [0.783, 0.887] & 0.821 [0.778, 0.861] \\
& & GPT-4o           & 0.874 [0.844, 0.903] & 0.813 [0.758, 0.865] & 0.873 [0.824, 0.920] & 0.842 [0.802, 0.879] \\
& & GPT-4o-Mini      & 0.870 [0.840, 0.899] & 0.817 [0.764, 0.869] & 0.852 [0.800, 0.901] & 0.834 [0.793, 0.873] \\
& & Llama-4-Maverick & 0.864 [0.833, 0.892] & 0.804 [0.747, 0.856] & 0.851 [0.798, 0.900] & 0.827 [0.784, 0.865] \\ 
\noalign{}
\midrule
\pagebreak
\setTK{Screening} 
\multirow{24}{*}{\rotv{Screening}} 
  \ptblockstart
& \multirow{6}{*}{\rotv{CoT}} & Claude-3.5-Haiku   & 0.824 [0.790, 0.858] & 0.789 [0.742, 0.834] & \textbf{0.912} [0.874, 0.946] & 0.846 [0.812, 0.877] \\
& & Deepseek-Chat-V3   & 0.838 [0.803, 0.870] & 0.866 [0.821, 0.907] & 0.819 [0.773, 0.865] & 0.842 [0.806, 0.874] \\
& & Gemini-2.5-Flash   & \textbf{0.874} [0.844, 0.903] & 0.869 [0.827, 0.907] & 0.896 [0.859, 0.932] & \textbf{0.883} [0.853, 0.910] \\
& & GPT-4o             & 0.872 [0.842, 0.901] & \textbf{0.889} [0.847, 0.926] & 0.865 [0.822, 0.906] & 0.877 [0.845, 0.906] \\
& & GPT-4o-Mini        & 0.856 [0.824, 0.886] & 0.880 [0.836, 0.919] & 0.842 [0.798, 0.886] & 0.861 [0.827, 0.891] \\
& & Llama-4-Maverick   & 0.848 [0.815, 0.878] & 0.834 [0.788, 0.875] & 0.888 [0.851, 0.926] & 0.860 [0.827, 0.890] \\
  \ptblocksep
  \ptblockstart
& \multirow{6}{*}{\rotv{CoT-few-shot}} & Claude-3.5-Haiku & 0.833 [0.799, 0.867] & 0.840 [0.795, 0.883] & 0.850 [0.806, 0.893] & 0.845 [0.811, 0.879] \\
& & Deepseek-Chat-V3 & 0.860 [0.830, 0.888] & 0.917 [0.879, 0.950] & 0.808 [0.759, 0.855] & 0.859 [0.825, 0.889] \\
& & Gemini-2.5-Flash & \textbf{0.878} [0.848, 0.907] & 0.882 [0.840, 0.918] & \textbf{0.888} [0.850, 0.925] & \textbf{0.885} [0.854, 0.913] \\
& & GPT-4o           & 0.870 [0.840, 0.899] & 0.912 [0.873, 0.946] & 0.835 [0.787, 0.876] & 0.871 [0.838, 0.901] \\
& & GPT-4o-Mini      & 0.846 [0.813, 0.876] & \textbf{0.922} [0.884, 0.955] & 0.773 [0.721, 0.824] & 0.841 [0.805, 0.875] \\
& & Llama-4-Maverick & 0.853 [0.821, 0.884] & 0.864 [0.820, 0.906] & 0.858 [0.814, 0.900] & 0.861 [0.827, 0.892] \\
  \ptblocksep
  \ptblockstart
& \multirow{6}{*}{\rotv{few-shot}} & Claude-3.5-Haiku & 0.849 [0.816, 0.880] & 0.847 [0.804, 0.889] & 0.873 [0.831, 0.912] & 0.860 [0.827, 0.890] \\
& & Deepseek-Chat-V3 & 0.870 [0.840, 0.901] & 0.902 [0.862, 0.936] & 0.846 [0.802, 0.891] & 0.873 [0.842, 0.903] \\
& & Gemini-2.5-Flash & \textbf{0.899 [0.870, 0.925]} & 0.895 [0.856, 0.930] & \textbf{0.915} [0.878, 0.948] & \textbf{0.905} [0.877, 0.930] \\
& & GPT-4o           & 0.897 [0.870, 0.923] & \textbf{0.930} [0.897, 0.960] & 0.869 [0.828, 0.909] & 0.899 [0.870, 0.925] \\
& & GPT-4o-Mini      & 0.862 [0.832, 0.892] & 0.910 [0.871, 0.944] & 0.819 [0.772, 0.864] & 0.862 [0.829, 0.893] \\
& & Llama-4-Maverick & 0.846 [0.813, 0.876] & 0.846 [0.801, 0.888] & 0.865 [0.824, 0.905] & 0.856 [0.822, 0.886] \\
  \ptblocksep
  \ptblockstart
& \multirow{6}{*}{\rotv{zero-shot}} & Claude-3.5-Haiku & 0.824 [0.790, 0.857] & 0.792 [0.746, 0.837] & 0.908 [0.870, 0.941] & 0.846 [0.812, 0.877] \\
& & Deepseek-Chat-V3 & 0.878 [0.850, 0.907] & 0.894 [0.855, 0.929] & 0.873 [0.833, 0.912] & 0.883 [0.853, 0.911] \\
& & Gemini-2.5-Flash & 0.872 [0.842, 0.901] & 0.843 [0.799, 0.883] & \textbf{0.931} [0.898, 0.960] & 0.885 [0.855, 0.911] \\
& & GPT-4o           & 0.874 [0.844, 0.903] & 0.887 [0.845, 0.923] & 0.873 [0.832, 0.912] & 0.880 [0.848, 0.908] \\
& & GPT-4o-Mini      & \textbf{0.890} [0.862, 0.917] & \textbf{0.933} [0.899, 0.962] & 0.854 [0.811, 0.895] & \textbf{0.892} [0.862, 0.919] \\
& & Llama-4-Maverick & 0.850 [0.817, 0.880] & 0.839 [0.794, 0.882] & 0.885 [0.846, 0.922] & 0.861 [0.829, 0.891] \\
\midrule
\setTK{Searching} 
\multirow{24}{*}{\rotv{Searching}} 
  \ptblockstart
& \multirow{6}{*}{\rotv{CoT}} & Claude-3.5-Haiku   & 0.841 [0.807, 0.873] & 0.652 [0.572, 0.726] & 0.828 [0.757, 0.891] & 0.729 [0.667, 0.785] \\
& & Deepseek-Chat-V3   & 0.870 [0.840, 0.899] & 0.774 [0.695, 0.849] & 0.701 [0.620, 0.778] & 0.736 [0.670, 0.793] \\
& & Gemini-2.5-Flash   & 0.878 [0.850, 0.907] & 0.748 [0.672, 0.819] & 0.795 [0.723, 0.863] & 0.771 [0.711, 0.825] \\
& & GPT-4o             & 0.884 [0.856, 0.913] & 0.754 [0.677, 0.824] & 0.819 [0.748, 0.884] & 0.785 [0.725, 0.835] \\
& & GPT-4o-Mini        & \textbf{0.923} [0.899, 0.945] & \textbf{0.803} [0.734, 0.865] & \textbf{0.929} [0.882, 0.970] & \textbf{0.861} [0.815, 0.902] \\
& & Llama-4-Maverick   & 0.860 [0.830, 0.888] & 0.746 [0.664, 0.821] & 0.693 [0.612, 0.770] & 0.718 [0.651, 0.779] \\
  \ptblocksep
  \ptblockstart
& \multirow{6}{*}{\rotv{CoT-few-shot}} & Claude-3.5-Haiku & 0.875 [0.845, 0.905] & 0.756 [0.678, 0.829] & 0.762 [0.685, 0.836] & 0.759 [0.696, 0.816] \\
& & Deepseek-Chat-V3 & 0.874 [0.844, 0.903] & 0.810 [0.733, 0.883] & 0.669 [0.583, 0.748] & 0.733 [0.664, 0.792] \\
& & Gemini-2.5-Flash & 0.870 [0.840, 0.899] & 0.733 [0.656, 0.804] & 0.780 [0.705, 0.850] & 0.756 [0.692, 0.811] \\
& & GPT-4o           & \textbf{0.905} [0.878, 0.929] & 0.828 [0.758, 0.892] & \textbf{0.795} [0.724, 0.863] & \textbf{0.811} [0.754, 0.860] \\
& & GPT-4o-Mini      & 0.899 [0.872, 0.925] & \textbf{0.841} [0.771, 0.905] & 0.748 [0.672, 0.822] & 0.792 [0.733, 0.845] \\
& & Llama-4-Maverick & 0.868 [0.837, 0.896] & 0.804 [0.726, 0.879] & 0.646 [0.558, 0.726] & 0.716 [0.645, 0.778] \\
   \ptblocksep
  \ptblockstart
 & \multirow{6}{*}{\rotv{few-shot}} & Claude-3.5-Haiku & 0.871 [0.841, 0.900] & 0.742 [0.667, 0.816] & 0.772 [0.698, 0.843] & 0.757 [0.696, 0.813] \\
& & Deepseek-Chat-V3 & 0.878 [0.848, 0.907] & 0.758 [0.680, 0.828] & 0.770 [0.695, 0.843] & 0.764 [0.700, 0.817] \\
& & Gemini-2.5-Flash & 0.880 [0.850, 0.909] & 0.721 [0.647, 0.791] & 0.874 [0.813, 0.929] & 0.790 [0.733, 0.839] \\
& & GPT-4o           & 0.862 [0.832, 0.892] & 0.675 [0.601, 0.742] & \textbf{0.898} [0.842, 0.947] & 0.770 [0.713, 0.820] \\
& & GPT-4o-Mini      & \textbf{0.903} [0.874, 0.929] & \textbf{0.776} [0.706, 0.843] & 0.874 [0.815, 0.929] & \textbf{0.822} [0.770, 0.870] \\
& & Llama-4-Maverick & 0.872 [0.841, 0.900] & 0.771 [0.694, 0.844] & 0.717 [0.636, 0.795] & 0.743 [0.677, 0.800] \\
 \ptblocksep
  \ptblockstart
& \multirow{6}{*}{\rotv{zero-shot}} & Claude-3.5-Haiku & 0.845 [0.812, 0.878] & 0.655 [0.580, 0.726] & 0.850 [0.789, 0.909] & 0.740 [0.681, 0.793] \\
& & Deepseek-Chat-V3 & 0.868 [0.838, 0.899] & 0.718 [0.641, 0.791] & 0.803 [0.733, 0.871] & 0.758 [0.698, 0.813] \\
& & Gemini-2.5-Flash & 0.864 [0.832, 0.895] & 0.690 [0.616, 0.760] & 0.858 [0.793, 0.916] & 0.765 [0.706, 0.815] \\
& & GPT-4o           & 0.860 [0.828, 0.890] & 0.671 [0.598, 0.739] & 0.898 [0.841, 0.948] & 0.768 [0.711, 0.818] \\
& & GPT-4o-Mini      & 0.872 [0.842, 0.901] & 0.688 [0.618, 0.757] & \textbf{0.921} [0.872, 0.964] & \textbf{0.788} [0.734, 0.836] \\
& & Llama-4-Maverick & \textbf{0.876} [0.846, 0.904] & \textbf{0.780} [0.705, 0.852] & 0.724 [0.644, 0.800] & 0.751 [0.687, 0.809] \\
\noalign{}
\pagebreak
\midrule
\setTK{Synthesis} 
\multirow{24}{*}{\rotv{Synthesis}} 
  \ptblockstart
& \multirow{6}{*}{\rotv{CoT}} & Claude-3.5-Haiku   & 0.839 [0.805, 0.870] & 0.616 [0.530, 0.700] & 0.733 [0.648, 0.815] & 0.670 [0.597, 0.736] \\
& & Deepseek-Chat-V3   & 0.840 [0.807, 0.872] & 0.627 [0.538, 0.716] & 0.679 [0.588, 0.768] & 0.652 [0.576, 0.723] \\
& & Gemini-2.5-Flash   & 0.805 [0.769, 0.840] & 0.547 [0.464, 0.628] & 0.697 [0.607, 0.781] & 0.613 [0.538, 0.680] \\
& & GPT-4o             & \textbf{0.858} [0.828, 0.888] & \textbf{0.676} [0.586, 0.758] & 0.688 [0.600, 0.772] & \textbf{0.682} [0.606, 0.749] \\
& & GPT-4o-Mini        & 0.838 [0.805, 0.870] & 0.604 [0.520, 0.684] & \textbf{0.771} [0.689, 0.846] & 0.677 [0.606, 0.742] \\
& & Llama-4-Maverick   & 0.832 [0.797, 0.866] & 0.607 [0.516, 0.694] & 0.679 [0.589, 0.766] & 0.641 [0.564, 0.710] \\
  \ptblocksep
  \ptblockstart
& \multirow{6}{*}{\rotv{CoT-few-shot}} & Claude-3.5-Haiku & 0.828 [0.794, 0.862] & 0.646 [0.539, 0.747] & 0.505 [0.408, 0.602] & 0.567 [0.478, 0.650] \\
& & Deepseek-Chat-V3 & 0.838 [0.803, 0.870] & 0.644 [0.547, 0.735] & 0.596 [0.500, 0.686] & 0.619 [0.536, 0.695] \\
& & Gemini-2.5-Flash & 0.826 [0.791, 0.858] & 0.592 [0.504, 0.675] & \textbf{0.679} [0.589, 0.764] & 0.632 [0.554, 0.703] \\
& & GPT-4o           & 0.852 [0.819, 0.884] & 0.676 [0.583, 0.765] & 0.633 [0.543, 0.724] & 0.654 [0.575, 0.724] \\
& & GPT-4o-Mini      & \textbf{0.874} [0.846, 0.903] & \textbf{0.783} [0.690, 0.870] & 0.596 [0.504, 0.688] & \textbf{0.677} [0.597, 0.750] \\
& & Llama-4-Maverick & 0.849 [0.819, 0.880] & 0.684 [0.589, 0.774] & 0.596 [0.505, 0.688] & 0.637 [0.556, 0.710] \\
  \ptblocksep
  \ptblockstart
& \multirow{6}{*}{\rotv{few-shot}} & Claude-3.5-Haiku & 0.833 [0.798, 0.865] & 0.651 [0.548, 0.750] & 0.519 [0.423, 0.614] & 0.577 [0.489, 0.657] \\
& & Deepseek-Chat-V3 & 0.842 [0.807, 0.874] & 0.645 [0.553, 0.733] & 0.633 [0.539, 0.724] & \textbf{0.639} [0.558, 0.711] \\
& & Gemini-2.5-Flash & \textbf{0.848} [0.815, 0.878] & \textbf{0.681} [0.581, 0.774] & 0.587 [0.495, 0.676] & 0.631 [0.549, 0.705] \\
& & GPT-4o           & 0.828 [0.793, 0.862] & 0.603 [0.513, 0.693] & 0.642 [0.550, 0.733] & 0.622 [0.545, 0.695] \\
& & GPT-4o-Mini      & 0.809 [0.775, 0.844] & 0.553 [0.468, 0.636] & \textbf{0.716} [0.627, 0.800] & 0.624 [0.549, 0.692] \\
& & Llama-4-Maverick & 0.837 [0.805, 0.870] & 0.649 [0.554, 0.743] & 0.578 [0.484, 0.669] & 0.612 [0.530, 0.683] \\
  \ptblocksep
  \ptblockstart
& \multirow{6}{*}{\rotv{zero-shot}} & Claude-3.5-Haiku & 0.841 [0.808, 0.873] & 0.621 [0.533, 0.704] & 0.713 [0.624, 0.797] & 0.664 [0.587, 0.730] \\
& & Deepseek-Chat-V3 & \textbf{0.846} [0.813, 0.878] & \textbf{0.639} [0.551, 0.725] & 0.697 [0.606, 0.781] & 0.667 [0.590, 0.734] \\
& & Gemini-2.5-Flash & 0.842 [0.809, 0.874] & 0.637 [0.544, 0.724] & 0.661 [0.571, 0.752] & 0.649 [0.570, 0.719] \\
& & GPT-4o           & \textbf{0.846} [0.813, 0.878] & 0.615 [0.535, 0.694] & \textbf{0.807} [0.730, 0.879] & \textbf{0.698} [0.629, 0.760] \\
& & GPT-4o-Mini      & 0.817 [0.783, 0.850] & 0.562 [0.484, 0.640] & 0.789 [0.707, 0.861] & 0.656 [0.586, 0.721] \\
& & Llama-4-Maverick & 0.827 [0.793, 0.860] & 0.595 [0.512, 0.680] & 0.688 [0.600, 0.773] & 0.638 [0.565, 0.706] \\

\midrule
\Needspace{100\baselineskip} 
\setTK{Writing} 
\multirow{24}{*}{\rotv{Writing}} 
  \ptblockstart
& \multirow{6}{*}{\rotv{CoT}} & Claude-3.5-Haiku   & 0.930 [0.907, 0.951] & 0.509 [0.380, 0.638] & 0.853 [0.722, 0.968] & 0.637 [0.513, 0.746] \\
& & Deepseek-Chat-V3   & 0.945 [0.925, 0.963] & 0.571 [0.423, 0.714] & 0.824 [0.692, 0.941] & 0.675 [0.543, 0.784] \\
& & Gemini-2.5-Flash   & 0.943 [0.921, 0.961] & 0.558 [0.414, 0.694] & 0.853 [0.719, 0.964] & 0.674 [0.545, 0.777] \\
& & GPT-4o             & 0.955 [0.937, 0.974] & 0.636 [0.487, 0.778] & 0.824 [0.679, 0.943] & 0.718 [0.588, 0.824] \\
& & GPT-4o-Mini        & \textbf{0.974} [0.959, 0.986] & \textbf{0.733} [0.596, 0.857] & \textbf{0.971} [0.897, 1.000] & \textbf{0.835} [0.732, 0.914] \\
& & Llama-4-Maverick   & 0.937 [0.915, 0.957] & 0.531 [0.383, 0.674] & 0.765 [0.613, 0.900] & 0.627 [0.492, 0.738] \\
  \ptblocksep
  \ptblockstart
& \multirow{6}{*}{\rotv{CoT-few-shot}} & Claude-3.5-Haiku & 0.958 [0.939, 0.975] & 0.760 [0.583, 0.917] & 0.576 [0.406, 0.743] & 0.655 [0.500, 0.786] \\
& & Deepseek-Chat-V3 & \textbf{0.970} [0.953, 0.984] & \textbf{0.806} [0.654, 0.933] & 0.735 [0.583, 0.885] & \textbf{0.769} [0.641, 0.871] \\
& & Gemini-2.5-Flash & 0.959 [0.941, 0.976] & 0.675 [0.523, 0.818] & \textbf{0.794} [0.643, 0.920] & 0.730 [0.597, 0.833] \\
& & GPT-4o           & 0.966 [0.947, 0.980] & 0.774 [0.615, 0.914] & 0.706 [0.548, 0.852] & 0.738 [0.600, 0.844] \\
& & GPT-4o-Mini      & 0.957 [0.939, 0.974] & 0.697 [0.533, 0.844] & 0.676 [0.516, 0.833] & 0.687 [0.542, 0.805] \\
& & Llama-4-Maverick & 0.957 [0.939, 0.974] & 0.667 [0.514, 0.815] & 0.765 [0.613, 0.900] & 0.712 [0.582, 0.822] \\
  \ptblocksep
  \ptblockstart
& \multirow{6}{*}{\rotv{few-shot}} & Claude-3.5-Haiku & 0.959 [0.941, 0.976] & \textbf{0.750} [0.581, 0.900] & 0.618 [0.448, 0.778] & 0.677 [0.526, 0.800] \\
& & Deepseek-Chat-V3 & 0.957 [0.939, 0.974] & 0.659 [0.510, 0.800] & 0.794 [0.655, 0.925] & 0.720 [0.590, 0.824] \\
& & Gemini-2.5-Flash & 0.959 [0.941, 0.976] & 0.694 [0.536, 0.842] & 0.735 [0.579, 0.875] & 0.714 [0.581, 0.821] \\
& & GPT-4o           & \textbf{0.963} [0.945, 0.980] & 0.700 [0.550, 0.838] & \textbf{0.824} [0.684, 0.939] & \textbf{0.757} [0.632, 0.854] \\
& & GPT-4o-Mini      & 0.941 [0.919, 0.961] & 0.551 [0.411, 0.694] & 0.794 [0.654, 0.920] & 0.651 [0.518, 0.762] \\
& & Llama-4-Maverick & 0.945 [0.925, 0.965] & 0.574 [0.426, 0.717] & 0.794 [0.654, 0.921] & 0.667 [0.533, 0.778] \\
  \ptblocksep
  \ptblockstart
& \multirow{6}{*}{\rotv{zero-shot}} & Claude-3.5-Haiku & 0.943 [0.922, 0.963] & 0.563 [0.425, 0.705] & 0.794 [0.645, 0.919] & 0.659 [0.531, 0.769] \\
& & Deepseek-Chat-V3 & \textbf{0.953} [0.933, 0.972] & 0.608 [0.468, 0.741] & 0.912 [0.806, 1.000] & \textbf{0.729} [0.609, 0.825] \\
& & Gemini-2.5-Flash & \textbf{0.953} [0.935, 0.972] & 0.612 [0.469, 0.750] & 0.882 [0.759, 0.974] & 0.723 [0.597, 0.822] \\
& & GPT-4o           & \textbf{0.953} [0.935, 0.972] & \textbf{0.622} [0.474, 0.761] & 0.824 [0.688, 0.941] & 0.709 [0.580, 0.811] \\
& & GPT-4o-Mini      & 0.941 [0.919, 0.961] & 0.542 [0.413, 0.672] & \textbf{0.941} [0.846, 1.000] & 0.688 [0.568, 0.788] \\
& & Llama-4-Maverick & 0.945 [0.925, 0.963] & 0.566 [0.429, 0.702] & 0.882 [0.767, 0.974] & 0.690 [0.568, 0.795] \\
\end{longtable}%
\vspace{-0.7\baselineskip} 
\begin{minipage}{\textwidth}
\small\emph{Note.} All metrics are reported with 95\% confidence intervals in brackets. 
Boldface highlights the best-performing \emph{point estimate} within each prompt type (under each task) and metric (CIs not bolded). 
The first two columns (TK, PT) are displayed vertically to save width. 
\textbf{TK} = Classification Tasks. \textbf{PT} = Prompt Type.
\end{minipage}

\section{Prompt Templates}\label{app:B}

\subsection{Zero-shot prompt template for SLR relevance classification}\label{app:B1}
\begin{promptbox}
{\fontsize{10}{13}\selectfont
You are analysing academic articles to determine whether they are relevant to the automation of the systematic literature review (SLR) process.\\[0.3em]
For each article, you are provided with the title and abstract. Based on this information, answer the following question:\\[0.3em]
Does the article discuss, study, or present methods related to the automation of systematic literature reviews or their component tasks?\\[0.6em]
\textbf{Definition of the SLR Process}\\[0.1em]
The systematic literature review (SLR) process includes the following five tasks:\\[-0.2em]
- Searching — retrieving relevant articles from databases or repositories\\
- Screening — selecting or filtering articles based on inclusion/exclusion criteria\\
- Retrieval — extracting information or data from article content\\
- Synthesis — summarizing, integrating, or organizing findings from selected studies\\
- Writing — generating narrative or structured text for the review\\[0.3em]
Automation methods may include algorithms, rule-based systems, traditional machine learning, or large language models (LLMs) such as GPT, BERT, etc.\\[0.6em]
\textbf{Inclusion Criteria}\\[0.1em]
Classify the article as **\textbf{relevant}** if **\textbf{any}** of the following apply:\\[-0.2em]
1. The article discusses or investigates the automation of systematic literature reviews or literature reviews, including conceptual discussions, tool development, system design, or frameworks, even if specific SLR tasks are not mentioned.\\
2. The article discusses, studies, or applies automation methods to perform **\textbf{any}** of the five defined SLR tasks — even if the article's primary purpose is not described as automating literature reviews — as long as the tasks are performed on academic or scientific literature.\\[0.6em]
\textbf{Exclusion Criteria}\\[0.1em]
The article is **\textbf{not relevant}** if:\\[-0.2em]
- It is a systematic review or literature review about automation applied in other domains (e.g., robotics, healthcare, smart factories), but does not address automation of literature reviews themselves.\\[0.6em]
\textbf{Response Format}\\[0.1em]
For each article, respond with: \texttt{"Yes"} if it meets the inclusion criteria, \texttt{"No"} if it matches the exclusion criteria or if there is insufficient information.\\[0.3em]
Now, analyse the following list of articles (each includes a title and abstract):\\[0.3em]
\texttt{\{title\_abstract\_pairs\}}\\[0.3em]
Return your analysis in JSON format, structured exactly like the following:\\[-0.2em]
[\\
\begin{verbatim}
    {
        "Title": "<article title>",
        "Relevant to SLR automation": "Yes" | "No"
    }
\end{verbatim}
]\\[0.2em]
Please strictly follow the JSON format, do not output any text other than the JSON objects. Make sure that double quotes inside value strings to be escaped as required by JSON.}
\end{promptbox}

\subsection{CoT prompt template for SLR relevance classification}\label{app:B2}

\begin{promptbox}
{\fontsize{10}{13}\selectfont
You are analysing academic articles to determine whether they are relevant to the automation of the systematic literature review (SLR) process.\\[0.3em]
Each article includes a title and abstract. Follow the reasoning steps below and then answer the final question.\\[0.6em]
\textbf{Step-by-Step Reasoning}\\[0.1em]
\textit{Step 1 - Identify Automation of Literature Review}\\
Is the article discussing, studying, or presenting methods related to automating literature reviews or systematic literature reviews?\\
The article is relevant **\textbf{as long as it discusses any aspect of automating literature reviews}** — including tool development, system architecture, conceptual frameworks, workflows, or general discussion — even if it does not specify which SLR task is automated.\\[0.3em]
\textit{Step 2 - Check for Automation of SLR Tasks}\\
Does the article describe or apply automation techniques to perform **\textbf{any}** of the following five tasks, specifically when applied to academic or scientific literature?\\[-0.2em]
- Searching — retrieving relevant articles from databases or repositories\\
- Screening — selecting or filtering articles based on inclusion/exclusion criteria\\
- Retrieval — extracting information or data from article content\\
- Synthesis — summarizing, integrating, or organizing findings from selected studies\\
- Writing — generating narrative or structured text for the review\\
The article is relevant \textbf{even if it does not mention literature reviews explicitly}, as long as it automates one or more of these tasks \textbf{on academic or scientific publications}.\\[0.3em]
\textit{Step 3 - Apply Exclusion Rules}\\
The article is \textbf{not relevant} if:\\[-0.2em]
- It is a literature review or systematic review of automation in a domain \textbf{unrelated to literature reviews} (e.g., automation in medicine, robotics, or manufacturing)\\
- It uses automation for tasks unrelated to literature review and does \textbf{not} apply methods to academic or scientific literature\\[0.3em]
\textit{Step 4 - Conclusion*}\\
Based on the steps above, decide whether the article is relevant to automation of literature reviews or automates any part of the SLR process.\\[0.6em]
\textbf{Final Question}\\[0.1em]
**Is the article relevant to the automation of systematic literature review (SLR) or any of its defined tasks?**
Respond with: \texttt{"Yes"} if it meets the inclusion criteria, \texttt{"No"} if it matches the exclusion criteria, or if there is insufficient information.\\[0.6em]
\textbf{Response Format}\\[0.1em]
Now, analyse the following list of articles. Each article includes a title and abstract:\\[0.3em]
\texttt{\{tittle and abstract pairs \}}\\[0.3em]
For each article, provide your reasoning and final decision in the following JSON format:\\[-0.2em]
[\\
\begin{verbatim}
    {{
        "Title": "<article title>",
        "Reasoning": "<brief explanation of your reasoning steps>",
        "Relevant to SLR automation": "Yes" | "No"
    }}
\end{verbatim}
]\\[0.2em]
Please strictly follow the JSON format, do not output any text other than the JSON objects. Make sure that double quotes inside value strings to be escaped as required by JSON.}
\end{promptbox}

\subsection{Zero-shot prompt template for SLR tasks classification}\label{app:B3}

\begin{promptbox}
{\fontsize{10}{13}\selectfont
You are analysing academic articles that have already been identified as relevant to the automation of systematic literature reviews (SLRs).\\[0.3em]
Each article includes a title and abstract. Your task is to determine:\\[0.3em]
\textit{1. Which SLR tasks are discussed or automated in the article?}\\
Select all applicable tasks from the list below:\\[-0.2em]
- Searching — fetching or retrieving candidate articles from databases, repositories, or search engines** (e.g., PubMed, Scopus)\\
- Screening — selecting or filtering articles from the search results, based on inclusion/exclusion criteria, relevance, or study characteristics\\
- Retrieval — extracting specific information or structured data from article content\\
- Synthesis — summarizing, integrating, or organizing findings across multiple studies\\
- Writing — generating narrative or structured text for the review\\[0.3em]
\textit{2. Does the article mention the use of large language models (LLMs)?}\\
Identify whether tools such as GPT, BERT, T5, or other LLMs are mentioned in the context of automating any of the above tasks.\\[0.6em]
\textbf{Response Format}\\[0.1em]
Respond clearly and concisely using the format below.\\[0.3em]
Now, analyse the following list of articles (each includes a title and abstract):\\[0.3em]
\texttt{\{title and abstract pairs\}}\\[0.3em]
Return your analysis in JSON format, structured exactly like the following:\\[-0.2em]
[\\
\begin{verbatim}
  {{
    "Title": "<article title>",
    "Searching automation": "Yes" | "No",
    "Screening automation": "Yes" | "No",
    "Retrieval automation": "Yes" | "No",
    "Synthesis automation": "Yes" | "No",
    "Writing automation": "Yes" | "No",
    "Using LLMs": "Yes" | "No"
  }}
\end{verbatim}
]\\[0.2em]
Please strictly follow the JSON format, do not output any text other than the JSON objects. Make sure that double quotes inside value strings to be escaped as required by JSON.}
\end{promptbox}

\subsection{CoT prompt template for SLR tasks classification}\label{app:B4}
\begin{promptbox}
{\fontsize{10}{13}\selectfont
You are analysing academic articles that have already been identified as relevant to the automation of systematic literature reviews (SLRs).\\[0.3em]
Each article includes a title and abstract. Your task is to determine:\\[0.3em]
\textit{1. Which SLR tasks are discussed or automated in the article?}\\
Select from the following tasks (choose all that apply):\\[-0.2em]
- Searching — fetching or retrieving candidate articles from databases, repositories, or search engines** (e.g., PubMed, Scopus)\\
- Screening — selecting or filtering articles from the search results, based on inclusion/exclusion criteria, relevance, or study characteristics\\
- Retrieval — extracting specific information or structured data from article content\\
- Synthesis — summarizing, integrating, or organizing findings across multiple studies\\
- Writing — generating narrative or structured text for the review\\[0.3em]
\textit{2. Does the article mention the use of large language models (LLMs)?}\\
Determine whether the article discusses LLMs such as GPT, BERT, T5, or similar, in the context of automating any part of the SLR process.\\[0.6em]
\textbf{Reasoning Instructions:}\\[-0.2em]
For each article, follow this reasoning process:\\[0.2em]
- First, briefly summarize what the article is about based on the title and abstract.\\
- Then, assess whether any of the five SLR tasks are described as being automated or supported.\\
- For each task, explain why you think it is (or is not) discussed.\\
- Finally, decide whether the article mentions the use of LLMs, citing keywords or model names if applicable.\\[0.6em]
\textbf{Response Format}\\[0.1em]
Then provide your structured output in the required JSON format.\\[0.3em]
Now, analyse the following list of articles. Each article includes a title and abstract:\\[0.3em]
\texttt{\{title and abstract pairs\}}\\[0.3em]
For each article, provide your reasoning and final decision in the following JSON format:\\[-0.2em]
[
\begin{verbatim}
  {{
    "Title": "<article title>",
    "Searching automation": "Yes" | "No",
    "Screening automation": "Yes" | "No",
    "Retrieval automation": "Yes" | "No",
    "Synthesis automation": "Yes" | "No",
    "Writing automation": "Yes" | "No",
    "Using LLMs": "Yes" | "No"
  }}
\end{verbatim}
]\\[0.2em]
Please strictly follow the JSON format, do not output any text other than the JSON objects. Make sure that double quotes inside value strings to be escaped as required by JSON.}
\end{promptbox}

\subsection{Self-reflection check in self-reflection prompt template for SLR relevance classification}\label{app:B5}

\begin{promptbox}
{\fontsize{10}{13}\selectfont
\textbf{Self-Reflection Check}\\[0.3em]
Before finalizing your decision, reflect on your reasoning:\\[0.2em]
- Did you mistakenly classify the article based only on keywords like "automation" or "systematic review" without considering the actual context?\\
- Did you overlook whether the automation applies to academic or scientific literature?\\
- Did you misclassify a literature review that focuses on automation in another domain (e.g., robotics, healthcare) as relevant, even though it is not about automating reviews themselves?\\
- Did you apply the inclusion and exclusion rules consistently?\\[0.3em]
If you find any mistakes or inconsistencies, revise your answer.}
\end{promptbox}

\subsection{System message}\label{app:B6}

\begin{promptbox}
{\fontsize{10}{13}\selectfont
You are an expert in machine learning, with specialized knowledge in large language models (LLMs) and the automation of systematic literature reviews (SLRs). You understand the structure of the SLR process and are able to identify whether and how automation techniques—particularly those involving LLMs—are applied to specific review tasks such as searching, screening, information extraction, synthesis, and writing.}
\end{promptbox}

\endgroup
\end{document}